%% file: neurips_2026.tex
\documentclass{article}

\usepackage[preprint]{neurips_2026}

\usepackage[utf8]{inputenc} % allow utf-8 input
\usepackage[T1]{fontenc}    % use 8-bit T1 fonts
\usepackage{hyperref}       % hyperlinks
\usepackage{url}            % simple URL typesetting
\usepackage{booktabs}       % professional-quality tables
\usepackage{amsfonts}       % blackboard math symbols
\usepackage{nicefrac}       % compact symbols for 1/2, etc.
\usepackage{microtype}      % microtypography
\usepackage{xcolor}         % colors
\usepackage{subcaption}
\usepackage[ruled,vlined]{algorithm2e}
\usepackage{graphicx}
\usepackage{caption}
\usepackage{threeparttable}
\usepackage{wrapfig}
\usepackage{pifont}
\usepackage{makecell}
\usepackage{hyperref}
\usepackage{pifont}
\usepackage{url}
\usepackage{enumitem}
\usepackage{booktabs}
\usepackage{multirow}
\usepackage{amsmath}
\usepackage{tikz}
\usepackage{listings}
\usepackage{soul}

\input{macros}

\title{\method: An Agentic Framework for Reaction Feasibility Prediction via Adaptive
Utility-aware Multi-tool Reasoning}

\author{Ye Liu$^{1}$, Botao Yu$^{2}$, Xinyi Ling$^{2}$, Daniel Adu-Ampratwum$^{3}$, Xia Ning$^{1,2,3,4}$ \\
$^1$Department of Biomedical Informatics \\
$^2$Department of Computer Science and Engineering \\
$^3$Division of Medicinal Chemistry and Pharmacognosy\\
$^4$Translational Data Analytics Institute \\
The Ohio State University \\
\texttt{\{liu.12989, ling.303, yu.3737, adu-ampratwum.1, ning.104\}@osu.edu} \\
}

\begin{document}

\maketitle

\begin{abstract}
Reaction feasibility prediction, 
as a fundamental problem in computational chemistry, 
has benefited from diverse tools enabled by recent advances in artificial intelligence, particularly large language models.
However, the performance of individual tools varies substantially across reactions,
making it difficult for any single tool to consistently perform well across all cases.
This raises a critical challenge: how to effectively leverage multiple tools to obtain more accurate feasibility predictions.
To address this, we propose \method, 
an agentic framework that explicitly models tool-specific utilities,
adaptively prioritizes tools, 
and further resolves the potential tool conflicts to produce the final prediction for each reaction.
Unlike existing approaches that rely on simple aggregation or heuristic assignment over various tools, 
\method organizes tools into a hierarchy 
that prioritizes top-performing tools and defers others when needed,
characterizes their strengths through tool-specific patterns, 
and resolves conflicts via memory-augmented reasoning.
Extensive experiments on a public dataset demonstrate that \method consistently outperforms strong baselines, 
including single-tool methods as well as various tool aggregation and tool selection approaches.
Further analysis shows that the improvements are particularly significant on reactions with conflicting tool predictions,
highlighting the effectiveness of \method in leveraging the complementary strengths of multiple tools.
%
% \ye{while also providing a transparent and interpretable decision process.}
%
The code is available via~\url{https://anonymous.4open.science/r/ARMOR-E13F}.
\end{abstract}

% \begin{abstract}
% Reaction feasibility prediction, 
% %
% as a fundamental problem in computational chemistry, 
% %
% has benefited from diverse tools enabled by recent advances in machine learning, particularly large language models.
% %
% Among various tools, no single tool consistently performs well across all reactions, 
% %
% as different tools exhibit complementary strengths on different reactions.
% %
% This raises a critical challenge: how to effectively leverage multiple tools to obtain more accurate feasibility prediction.
% %
% To address this, we propose \method, 
% % 
% an agent-based framework that 
% % 
% models the capabilities of different tools and leverages these capabilities to select appropriate tools for producing the final prediction for each reaction.
% % 
% Unlike existing approaches that rely on simple aggregation or heuristic assignment, 
% % 
% \method organizes tools hierarchically to determine the usage of different tools. 
% % 
% And then profiles tool strengths via capability patterns, 
% % 
% and resolves conflicts through memory-augmented reasoning.
% %
% Extensive experiments on the publicly available dataset demonstrate that \method consistently outperforms strong baselines, 
% %
% including single-tool methods as well as various tool aggregation and tool selection approaches.
% %
% Further analysis reveals that the improvements are particularly significant on reactions with conflicting tool predictions, 
% %
% while \method provides a transparent and interpretable tool selection process.
% \end{abstract}

%%%%%%%%%%%%%%%%%%%%%%%%%%%%%%%
\section{Introduction}
\label{intro}
%%%%%%%%%%%%%%%%%%%%%%%%%%%%%%%

Reaction feasibility prediction, 
which aims to assess whether a chemical reaction is feasible or not,  
is a fundamental problem in computational chemistry and chemical synthesis~\citep{warr2014short}. 
Recent advances in artificial intelligence,
particularly large language models (LLMs), 
have led to a variety of tools for this task,  
such as classification-based feasibility predictors~\citep{chainani2025dora}, 
forward-generative models~\citep{irwin2022chemformer}, 
and LLM-based feasibility reasoners~\citep{rubin2022learning}.
However, the performance of existing tools varies across reactions, and no single tool consistently produces correct predictions in all cases.
For example, classification-based predictors tend to perform well on reactions with relatively regular structures~\citep{probst2022reaction}, 
while LLM-based methods are more effective on reactions that require complex reasoning or contextual understanding~\citep{krishnan2026biomodelsrag}.
Such complementary strengths suggest that effectively leveraging multiple tools for each reaction is crucial for more accurate feasibility prediction.

To this end,
prior work has explored leveraging multiple tools, 
such as dynamic ensemble selection methods~\citep{cruz2020deslib} and mixture-of-experts models~\citep{huang2024harder}.
However, these approaches typically rely on simple aggregation or heuristic assignment strategies,
without explicitly distinguishing when each tool is more appropriate for which reaction, 
making their final predictions less stable or accurate.
Meanwhile, recent advances in LLM-based agents have demonstrated strong capability in coordinating multiple tools through planning and reasoning~\citep{qu2025tool,qintoolllm,ye2025tooleyes}.
Nevertheless, they mainly focus on orchestrating tools with different functionalities to accomplish multi-step tasks,
where each tool typically serves a distinct sub-goal.
In contrast, leveraging multiple tools for a single task,
especially in domain-specific scenarios such as reaction feasibility prediction,
remains underexplored.

In this paper,
we study how to effectively leverage multiple tools for accurate reaction feasibility prediction,
where different tools exhibit varying strengths across reactions.
We propose \method,
an \textbf{A}gentic framework for adaptive utility-aware \textbf{M}ulti-tool \textbf{R}easoning and c\textbf{O}nflict resolution for \textbf{R}eaction feasibility prediction, which
%
%
%an agentic framework that 
% 
models tool utilities, prioritize tools with respect to different reactions
and further resolves the potential tool conflicts with the support of contrastive demonstrations.
%
%Specifically, 
\method consists of three key components:
\textbf{(1)} a tool hierarchy construction module that organizes multiple tools into a two-level structure, where the first level includes top-performing tools for initial decision-making, while the second level contains remaining tools that exhibit specialized performance for reactions with specific characteristics;
\textbf{(2)} a utility-aware tool prioritization module that selects tools by characterizing the reaction-specific utilities of different tools, thus adaptively prioritizing tools that tend to make correct predictions for each reaction.
and \textbf{(3)} a tool conflict resolution module that resolves the potential conflicting predictions of selected tools via a novel memory-augmented reasoning mechanism, which leverages the historical reasoning behaviors over contrastive reaction-tool demonstrations to obtain the final predictions.
Overall, \method enables reaction-specific utility assessment over multiple tools,
adaptively prioritizes appropriate tools and resolves tool conflicts to accurately predict reaction feasibility.
% \hl{provides a systematic agentic framework to leverage multiple tools with varying performance,
% % 
% by explicitly assessing each tool's utility, adaptively selecting appropriate tools and reconciling tool conflicts,
% % 
% to derive an accurate feasibility prediction for each reaction.} \xia{looks repeated for the first 
% opening sentence; rephrase}

Although we focus on reaction feasibility prediction in this work,
the proposed \method is a general framework that can be applied to other tasks where different tools exhibit varying performance across inputs.
We use the reaction feasibility prediction task as a representative setting to evaluate \method, 
%the proposed framework,
%
and conduct extensive experiments on the public reaction feasibility dataset with diverse tools.
The results demonstrate that \method consistently outperforms strong baselines, including single-tool methods and various tool aggregation and tool selection approaches.
In particular, \method achieves superior and balanced performance,  
without bias toward either feasible or infeasible reactions. 
% \hl{while maintaining a better balance across both feasible and infeasible reactions.} \xia{what does this mean? }
%
Further analysis shows that the performance gain primarily stems from its ability to model tool-specific utilities and effectively resolve tool conflicts. 
%\ye{how about directly deleting the following sentence?}
%\hl{thereby validating the design of the proposed framework.} \xia{awkward language.. there is no ``validation'' of a model design... }

Our contributions can be summarized as follows: 
% 
% \begin{itemize}[leftmargin=12pt, itemsep=0pt, topsep=0pt] 
    % 
    % \item 
    (1) We investigate how to leverage multiple tools for reaction feasibility prediction and, for the first time,
    propose the direction of explicit modeling the utilities of different tools, 
    moving beyond direct aggregation or heuristic mixture-of-expert approaches.
    % 
    % \item 
    (2) We develop \method, an agentic framework 
    that characterizes tool utilities and leverages them to select appropriate tools for each reaction, followed by a tool conflict resolution module to derive the final predictions.
    % 
    % \item 
    (3) Extensive experiments demonstrate that \method consistently achieves state-of-the-art performance.
    % across various metrics. 
    % 
    It shows significant advantages on reactions with persistent tool conflicts,
    where leveraging complementary tool strengths becomes especially important.
    % it provides a \hl{transparent and interpretable} \xia{is this emphasized as 
    % inspiration, motivation, or rationale in method? } decision process. 
    % 
    % 
% \end{itemize}

% \begin{wrapfigure}{r}{0.55\textwidth}
%   \centering
%   \includegraphics[width=0.525\textwidth]{figures/motivation.pdf}
%   \caption{Example figure.}
%   \label{fig:motivation}
% \end{wrapfigure} 
% 

% Generally, reaction feasibility prediction can be formulated as a binary classification problem~\citep{yang2024subgraph}.  
% % 
% Earlier rule-based or heuristic systems assess feasibility through handcrafted chemical constraints~\citep{jorgensen1990cameo},  
% % 
% while the following deep learning approaches improve predictive accuracy using reaction fingerprints, graph neural networks~\citep{fooshee2018deep}.  
% % 
% More recently,  
% % 
% LLM-based methods exhibit strong chemical understanding, showing promising abilities in mechanistic explanation,  
% % 
% reaction analysis, and knowledge-driven reasoning~\citep{murakumo2023llm}. 

% Despite their effectiveness,  
% 

\section{Related Work}

\paragraph{Reaction Feasibility Prediction}
Early studies on reaction feasibility prediction relied on expert-designed rules~\citep{warr2014short,zhong2025towards}, 
which assess feasibility through handcrafted constraints such as functional-group compatibility and valence rules~\citep{jorgensen1990cameo,aithal2012feasibility}. 
While these methods provide interpretable decision criteria, they require substantial manual effort and often fail to generalize beyond predefined rule sets~\citep{fooshee2018deep}.
To overcome these limitations, 
subsequent work has explored machine learning approaches, 
which can be broadly categorized into classification-based predictors and forward-generation models~\citep{park2022machine}. 
Classification-based methods train supervised models using reaction fingerprints or molecular descriptors~\citep{probst2022reaction,yang2024subgraph,chainani2025dora}, 
while forward-generation models assess feasibility by predicting plausible products and checking their consistency with the target outputs~\citep{schwaller2019molecular,irwin2022chemformer}. 

More recently, 
large language models (LLMs) have been applied to reaction understanding and synthesis planning~\citep{murakumo2023llm}. 
For feasibility prediction, LLMs can perform zero-shot reasoning or leverage in-context learning (ICL)~\citep{kojima2022large,brown2020language}, 
and can be further enhanced by incorporating additional, external signals
% \hl{such as retrieved examples or model outputs} 
% \xia{make it more concrete within the context of feasibility prediction}
%\ye{making it more concrete within the context of feasibility prediction}
~\citep{rubin2022learning,krishnan2026biomodelsrag}. 
These advances highlight the flexibility of LLMs in 
integrating diverse information and performing contextual reasoning
% \hl{integrating diverse information} \xia{is this 
% the key? just wanted to double check} 
for reaction analysis.

\paragraph{Tool Selection} 
%\ye{I checked it. Your feeling is correct. I'd like to choose one from tool selection and multi-tool selection. What do you think?}
% 
Selecting appropriate tools for a given input has been widely studied in machine learning and AI systems~\citep{cruz2020deslib}. 
Early approaches rely on static strategies such as majority voting or weighted aggregation~\citep{dietterich2000ensemble}, 
which combine predictions without considering instance-specific characteristics and thus fail to fully exploit tool complementarity~\citep{rokach2010ensemble}.
To address this, 
dynamic selection methods have been proposed to adaptively select tools based on input characteristics~\citep{britto2014dynamic}. 
Representative approaches include dynamic ensemble selection methods such as KNORA~\citep{ko2008dynamic} and DES variants~\citep{woloszynski2012measure}, 
as well as mixture-of-experts (MoE) models that route inputs to different experts via learned gating mechanisms~\citep{shazeer2017outrageously,huang2024harder}. 
However, these methods typically rely on implicit competence estimation. 
% and 
%\ye{make sense. I think we also do not have experiments to demonstrate it. I will remove such expressions.}
%\hl{often lack interpretability.} \xia{I think at least MoE has interpretability. In addition, if we make the 
%contraction on interpretability, we should highlight interpretablity of our method, which is not done yet
%in Method...}

Beyond these approaches %\xia{in this paragraph, we need to mention the word ``agent" or ``agentic"}
and benefiting from the advances of LLMs, 
a variety of agent-based tool selection methods have been explored~\citep{qin2024tool,qu2025tool}, 
where the LLM-based agents are used to coordinate multiple tools for complex tasks. 
In such settings, tools are usually assigned to different sub-tasks, % within a pipeline, 
and the agent focuses on orchestrating their interactions~\citep{qintoolllm}. 
In contrast, leveraging multiple tools for a single task, 
remains underexplored,
especially in domain-specific scenarios, such as reaction feasibility prediction, where multiple tools with the same functionalities need be compared. 
Our work addresses this gap by explicitly modeling tool utilities and adaptively selecting appropriate tools for each reaction.

\section{\method Framework}
\label{sec:method}

\paragraph{Problem Definition}
\label{subsec:problem}

%Reaction feasibility prediction is formulated as a binary classification problem, 
%%
%where each reaction is classified as either feasible ($y = 1$) or infeasible ($y = 0$)~\citep{yang2024subgraph}. 
%%
%Formally, given a reaction $r = (S \rightarrow P)$, where $S$ and $P$ denote the reactants and products,  respectively, 
%%
%the goal is to predict its feasibility label.
%In this work, we consider a setting with multiple feasibility prediction tools 
%$\mathcal{T}^{(\text{all})} = \{t_i\}$, 
%%
%where each tool $t_i$ produces a prediction $f_{i}(r) \in \{0,1\}$ for a given reaction $r$. 
%%
%Our goal is to leverage multiple tools to derive the final prediction through an agentic framework.
%%
%% The key challenge is to identify which tools are more suitable for a given reaction 
%% and how to use their predictions to obtain the final decision.

Given a reaction $r = (S \rightarrow D)$, where $S$ and $D$
% \xia{$P$ is used later for patterns; need to change here} 
denote the reactants and products,  respectively,
reaction feasibility prediction is formulated as a binary classification problem to 
predict $r$ as either feasible ($y = 1$) or infeasible ($y = 0$)~\citep{yang2024subgraph}. 
In this work, we consider a setting in which multiple feasibility prediction tools
$\mathcal{T} = \{t_i\}$ are available and have varying performance across reactions, and 
we aim to optimally leverage these tools to derive accurate predictions through a novel agentic framework, \method.

\paragraph{Overview}

\begin{figure*}[t]
    \centering
%    \vspace{-15pt}
    \includegraphics[width=0.95\textwidth]{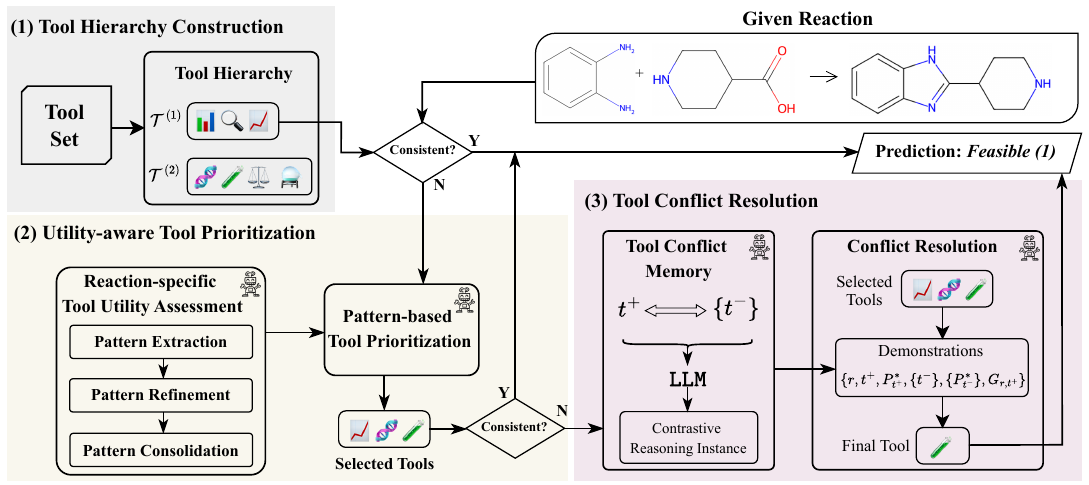}
    \caption{\method framework. The robot icon indicates that the corresponding module is agentic.}
    \label{fig:framework}
    \vspace{-15pt}
\end{figure*}

%\method is an agentic framework for reaction feasibility prediction that leverages multiple tools by modeling their respective capabilities and resolving their conflicts.
\method dynamically identifies the most suitable tool(s) for reaction feasibility prediction by 
measuring tool utilities and prioritizing tools with respect to different reactions, and resolving 
potential prediction conflicts from different tools by learning and reasoning through contrastive demonstrations. 
As illustrated in Figure~\ref{fig:framework},
\method consists of three key components:
%
%(1) Tool Hierarchy Construction (Section~\ref{sec:tool_hierarchy}), 
%% 
%which organizes tools into two levels, where the first level contains the top-performing tools for initial decision-making, 
%% 
%while the second level includes the remaining tools, which are used for cases that level-1 tools cannot address;
%
\textbf{(1)} Tool Hierarchy Construction (Section~\ref{sec:tool_hierarchy}), 
which organizes tools into two levels: 
tools on the first level have strong overall performance across reactions, and can therefore be used for initial 
prediction, whereas 
tools on the second level exhibit more pronounced reaction-dependent performance 
and thus, are better suited for specialized prediction scenarios or reactions with specific characteristics. 
\method leverages such a hierarchical tool framework to balance overall robustness with reaction-specific specialization in tool utilization. 
%	
%where the first level includes the top-performing tools for initial decision-making, 
%% 
%while the second level includes the remaining tools, which are used for cases that level-1 tools cannot address;
%
%\textbf{(2)} Capability-aware Tool Selection (Section~\ref{subsec:cap_tool_sel}), 
%% 
%which performs tool selection by leveraging patterns that characterize tool-specific capabilities, 
%% 
%focusing on tools that are more likely to make correct predictions;
%
\textbf{(2)} Utility-aware Tool Prioritization (Section~\ref{subsec:cap_tool_sel}), 
% 
%which conducts tool selection by leveraging patterns that characterize tool utilities over different reactions, 
%% 
%prioritizing different tools that are more likely to make correct predictions for different reactions;
which performs tool selection by leveraging patterns that characterize tool utility across different reactions, 
thereby prioritizing tools that are more likely to generate correct predictions for specific reaction characteristics.
This prioritization enables more adaptive and reaction-aware tool utilization, 
improving predictive accuracy while better leveraging the complementary strengths of different tools.
\textbf{(3)} Tool Conflict Resolution (Section~\ref{subsec:conflict}), 
which reconciles conflicting predictions from selected tools 
through novel memory-augmented reasoning over contrastive reaction-tool demonstrations 
to derive the final prediction.
The conflict resolution enables more reliable and context-aware decision-making 
by leveraging complementary evidence and historical reasoning behaviors
across tools.
% 
%which resolves the conflicts among selected tools through a novel memory-augmented reasoning 
%% 
%to determine the final prediction. 

%===============================================================
\subsection{Tool Hierarchy Construction}
\label{sec:tool_hierarchy}
%===============================================================

It is commonly observed that tool performance often varies across reactions, 
with no single tool consistently performing best in all cases. 
%
%Different tools exhibit varying performance across reactions, 
%% 
%and no single tool performs best in all cases.
%
%To optimally leverage these tools, 
To assess overall performance, 
% 
%\method ranks all tools in descending order according to their performance on the validation set with ground-truth labels.
\method quantifies the performance of the tool set $\mathcal{T}$ on the validation set (e.g., using accuracy),
% \xia{what metrics?}), 
with the top $\rho\%$ best-performing tools categorized into the first level of 
the tool hierarchy, denoted as $\mathcal{T}^{(1)}$, and 
the remaining tools in the second level, denoted as $\mathcal{T}^{(2)}$, 
% 
%The tool set $\mathcal{T}^{(\text{all})}$ is subsequently partitioned into a two-level hierarchy:
%
%the top $\rho\%$ of tools forms the first level $\mathcal{T}^{(1)}$,
%%
%while the remaining tools form the second level $\mathcal{T}^{(2)}$,
%
where $\rho \in (0,100)$ is a predefined ratio.
%
%Level-1 tools ($\mathcal{T}^{(1)}$) exhibit generally good performance
$\mathcal{T}^{(1)}$ exhibit strong overall performance
and thus are utilized for the initial decision-making,
while $\mathcal{T}^{(2)}$ provide complementary evidence and will be used when
reactions cannot be consistently handled by $\mathcal{T}^{(1)}$.
This hierarchical organization enables \method to adaptively leverage tools for different reactions, 
leading to more accurate predictions.

% \paragraph{Level-1 Conflict Identification.}
% Given the constructed tool hierarchy ($\mathcal{T}^{(1)}, \mathcal{T}^{(2)}$),
% %
% reactions in the validation set can be naturally partitioned based on the predictions of level-1 tools ($\mathcal{T}^{(1)}$).
% %
% Reactions on which level-1 tools produce consistent predictions
% %
% can be resolved with high confidence,
% %
% while those with inconsistent predictions are identified as level-1 conflict reactions.
% %
% In particular,
% %
% we treat such inconsistencies as signals,
% %
% indicating that different tools capture complementary aspects of the reaction.
% %
% These reactions often correspond to complex or ambiguous scenarios,
% %
% and are therefore further analyzed
% %
% to characterize tool-specific capabilities
% %
% and support instance-level tool selection.

\label{subsec:tool_profiling}

\vspace{-5pt}
%========================================================
\subsection{Utility-aware Tool Prioritization}
\label{subsec:cap_tool_sel}
%========================================================

%Given a reaction $r$,
%%
%\method first applies level-1 tools ($\mathcal{T}^{(1)}$) to produce initial predictions for $r$.
%%
%If the initial predictions are consistent,
%%
%\method directly outputs the consensus as the final prediction.
%%
%Otherwise,
%%
%$r$ is considered more challenging,
%%
%indicating that relying solely on level-1 tools is insufficient.
%%
%In such cases,
%% 
%\method proposes to model the capabilities of each tool in $\mathcal{T}^{(\text{all})}$, and 
%% 
%leverages these capabilities to select appropriate tools for each reaction.

Given a reaction $r$, \method first applies tools of $\mathcal{T}^{(1)}$ to produce initial predictions. 
If these predictions are consistent, indicating strong agreement and high confidence among tools 
with robust overall performance, \method adopts the consensus as the final prediction.
Otherwise, discrepancies among predictions from \levelOne are further addressed through 
utility-aware tool prioritization, which selectively leverages tools from both \levelOne and \levelTwo 
based on their reaction-specific utilities.
In order to select the most suitable tool for the reaction $r$, \method utilizes a two-step process: 
\textbf{(1)} Assess tool utilities using patterns; and 
\textbf{(2)} select tools using their patterns with respect to $r$.  

%===============================================
\subsubsection{Reaction-specific Tool Utility Assessment}
%===============================================

%----------------------------------------------------------------------------------
\paragraph{Pattern Extraction}
\label{subsubsec:pattern_extraction}
%----------------------------------------------------------------------------------
% Double Bond Formation
% The tool correctly predicts reactions where double bonds are formed between atoms that were previously single-bonded. This includes cases where a double bond is formed between two carbon atoms, or between a carbon atom and another atom like nitrogen or oxygen.

\method uses discrete, descriptive patterns to summarize the situations when each tool is 
likely to make correct reaction feasibility predictions, and thus, to characterize tool utilities. 
Specifically, a tool-specific pattern is defined as a tuple: $P_t = (d_t, e_t, \mathcal{X}_t)$,
where $d_t$ is a succinct description (e.g., ``double bond formation"), 
$e_t$ is a textual explanation of the conditions under which the tool $t$ performs well 
(e.g., ``The tool correctly predicts reactions where double bonds are formed between atoms that were previously single-bonded. This includes cases where a double bond is formed between two carbon atoms, or between a carbon atom and another atom like nitrogen or oxygen"), 
and $\mathcal{X}_t$ is a set of representative reaction examples covered by this pattern.
Such patterns are extracted from reactions in the validation set for which \levelOne fails to 
produce consistent predictions -- this set of reactions is denoted as \Rset. 
The reactions in \Rset are particularly informative for characterizing tool utilities, 
as consistently predicted reactions provide limited discriminative signals 
for differentiating the complementary strengths and weaknesses of individual tools. 

To extract $\mathcal{X}_t$ from \Rset, \method constructs $M$ diagnostic subsets for tool $t$ from \Rset,
denoted as $\{\mathcal{R}_t^{(m)}\}_{m=1}^{M}$, 
with each subset $\mathcal{R}_t^{(m)} = \{\{r_{11}\}, \{r_{10}\}, \{r_{01}\}, \{r_{00}\}\}_t$, 
%includes four types of cases with respect to $t$,  
where $\{r_{xy}\}$ represents a set of $N$ reactions with ground-truth feasibility label $x$ and $t$'s predicted feasibilty label $y$ ($1/0$ representing feasible/infeasible).  
These diagnostic subsets provide a comprehensive basis for analyzing $t$'s behavior over 
informative and discriminative reactions.
Meanwhile, using subsets instead of the entire \Rset improves the efficiency in extracting many patterns.

\method subsequently extracts patterns from $\mathcal{R}_t^{(m)}$ using a large language model $\mathtt{LLM}$, 
and aggregates the extracted patterns across all $M$ diagnostic subsets to obtain the pattern set 
\PSZero for tool $t$, as follows:
\begin{equation}
\label{eq:extraction}
    \PSZero = \bigcup\nolimits_{m=1}^{M} \mathtt{LLM}(\mathcal{R}_t^{(m)}). 
\end{equation}
\paragraph{Pattern Refinement}
%----------------------------------------------------------------------------------

% 
%To ensure the utility of \PSZero for tool $t$, \method validates each pattern $P_{tj} = (d_{tj}, e_{tj}, \mathcal{X}_j) \in \PSZero$ 
%from two perspectives: 
$\mathtt{LLM}$ may make mistakes when extracting patterns. 
To eliminate those mistakenly/inaccurately generated patterns 
by $\mathtt{LLM}$, \method further refines 
\PSZero from two perspectives:  
\textbf{(1)} How well pattern $P_{tj} = (d_{tj}, e_{tj}, \mathcal{X}_{tj}) \in \PSZero$ truly reflects the behavior of tool $t$,
using a score $\texttt{Align}(P_{tj})$,
defined as the proportion of reactions in $\mathcal{X}_{tj}$ that are correctly predicted by $t$.
Higher $\texttt{Align}(P_{tj})$ indicates that $P_{tj}$ better reflects the utility of $t$.
\textbf{(2)} How well pattern $P_{tj}$ covers its representative reaction examples $\mathcal{X}_{tj}$, 
using a score $\texttt{Cov}(P_{tj})$, defined as the proportion of reactions in $\mathcal{X}_{tj}$ 
that are covered by $P_{tj}$. 
Higher values indicate better coverage.
%
%
%
%
%% 
%To ensure the utility of the detected pattern set $\mathring{\mathcal{P}}_i$ for $t_i$, \method validates each pattern $P_{ij} \in \mathring{\mathcal{P}}_i$ from two perspectives: 
%%
%(1) whether pattern $P_{ij}$ aligns with the behavior of tool $t_i$, and 
%%
%(2) whether pattern $P_{ij}$ consistently covers its representative reaction examples $\mathcal{X}_{ij}$.
%%
%
%\begin{itemize}[leftmargin=12pt, itemsep=0pt, topsep=0pt]
%
%\item \textbf{Behavior Alignment.}
%It quantifies the alignment between pattern $P_{ij}$ and the behavior of its associated tool $t_i$,
%%
%using $\texttt{Align}(P_{ij})$,
%%
%defined as the proportion of reactions in $\mathcal{X}_{ij}$ that are correctly predicted by $t_i$.
%%
%Higher $\texttt{Align}(P_{ij})$ indicates that $P_{ij}$ better reflects the capabilities of $t_i$.
%
%\item \textbf{Pattern Coverage.}
%%
%It quantifies how consistently pattern $P_{ij}$ covers its representative reaction examples $\mathcal{X}_{ij}$.
%%
%To this end,
%%
%\method defines $\texttt{Cov}(P_{ij})$ as the proportion of reactions in $\mathcal{X}_{ij}$ that are covered by $P_{ij}$, as determined by the $\mathtt{LLM}$.
%%
%Higher values indicate better coverage.
%%
%\end{itemize}
%
%Finally,
%
Thus, \method retains only patterns that satisfy the following:
\begin{equation}
\label{eq:p_valid}
    \PSOne = \{\, P_{tj} \mid \texttt{Align}(P_{tj}) \ge \tau_1,\; \texttt{Cov}(P_{tj}) \ge \tau_2 \,\},
\end{equation}
where $\tau_1, \tau_2 \in [0,1]$ are the thresholds to control the quality of the retained pattern set 
\PSOne.

% 
%----------------------------------------------------------------------------------
\paragraph{Pattern Consolidation}
%----------------------------------------------------------------------------------

It is possible that in \PSOne, 
different patterns extracted from different reactions describe similar tool utilities. 
%as similar conditions can be identified from different reactions.
%
To obtain a representative, compact, and non-redundant pattern set,
\method consolidates patterns sharing the same succinct descriptions, and only retains one representative pattern 
out of them, selected by $\mathtt{LLM}$. 
For each pattern $P_{tj}$ remaining in this consolidated pattern set, \method further measures its quality on the entire \Rset (note that 
all the patterns are extracted from subsets of \Rset), using a confidence score defined as follows: 
\begin{equation}
\label{eq:conf}
\texttt{Conf}(P_{tj}) = 
\frac{
\left| \{\, r \in \Rset \mid r \text{ is covered by } P_{tj},\; f_{t}(r) \text{ is correct} \,\} \right|
}{
\left| \{\, r \in \Rset \mid r \text{ is covered by } P_{tj} \,\} \right|
},
\end{equation}
where $f_{t}(r)$ is the predicted feasibility of $r$ by tool $t$ associated with $P_{tj}$. 
Compared to $\texttt{Align}(P_{tj})$, which is measured on $\mathcal{X}_{tj}$ that are subsets of \Rset, 
$\texttt{Conf}(P_{tj})$ offers a more holistical assessment over the entire \Rset, and thus, enabling a more reliable 
tool selection based on the consolidated pattern set. 
For each tool $t$ in this consolidated pattern set, 
\method retains its top-5 patterns with $\texttt{Conf}_{tj}$ score above $\tau_3$ ($\tau_3 \in [0,1]$), 
and thus, constructs the final pattern set, denoted as \PSfinal, that will be used for tool prioritization and conflict resolution.

\subsubsection{Pattern-based Tool Selection}
\label{sec:pattern_matching}
%==========================================================

To accurately predict reaction feasibility of a new reaction $r'$ during inference time, \method first applies \levelOne tools. 
If no consensus from \levelOne tools, \method proceeds to select the most suitable tools from both \levelOne and \levelTwo, 
based on $r'$-specific tool utilities captured in $\{\PSfinal\}$. 
Specifically, \method first identifies the patterns in $\{\PSfinal\}$ that cover $r'$. 
From those covering patterns, \method identifies their associated tools, 
and the top-1 most confident pattern, in terms of $\texttt{Conf}$ scores, of each tool $t$, that covers $r'$ 
(The tools may have patterns not covering $r'$). This pattern is denoted as $P^*_t(r')$. 
Based on $P^*_t(r')$'s $\texttt{Conf}$ scores, the top-$L$ tools are selected to predict $r'$. This set of selected tools is denoted as $\SelectedTool(r')$. 
This tool selection, grounded on historical patterns and tool performance over the patterns, 
%%=======
%This tool selection based on historical
%\ye{I have one concern about the word "historical", which may be confused with the historical reasoning behaviors we design in conflict resolution}
%patterns and tool performance over the patterns 
%%>>>>>>> bb862c2b44aba3d85b7c382967fa555269ee5ada
enables \method to identify reaction-specific tool strengths, leading to more accurate predictions.
Meanwhile, leveraging multiple tools provides opportunities for \method to consider diverse patterns covering the reaction, 
leading to more robust predictions. 
If these tools produce consistent predictions for $r'$,
\method outputs the consensus as the final prediction.
Otherwise, it proceeds to the conflict resolution stage. 
\subsection{Tool Conflict Resolution}
\label{subsec:conflict}
%=========================================================

\paragraph{Tool Conflict Memory}

\method learns to resolve conflicts among tool predictions via a novel tool conflict memory \memory. 
To construct \memory, \method constructs structured contrastive instances \CTLI for each reaction $r\in\Rset$ as follows: 
\begin{equation}
\CTLI_{r, t^+} = \{r, t^+, P^*_{t^+}(r), \{t^-\}, \{P^*_{t^-}(r)\}\}, 
\end{equation}
where $t^+$ is a tool that accurately predicts $r$, with $P^*_{t^+}(r)$ its most confident pattern covering $r$; 
$\{t^-\}$ is a set of tools that fail to accurately predict $r$, with $\{P^*_{t^-}(r)\}$ the set of most confident patterns of $\{t^-\}$
covering $r$. 
The $\mathtt{LLM}$ is employed to generate a rationale $G_{r, t^+}$ explaining why $t_i^+$ is more suitable for reaction $r$ than $\{t_j^-\}$ based on their associated patterns.
$G_{r,t^+}$ is further incorporated into $\CTLI_{r,t^+}$,
yielding the final contrastive instance stored in $\memory$.
These contrastive instances highlight the performance differences among tools on the same reaction, as well as their patterns inducing 
tool predictions.
The memory $\mathcal{M}$ stores such instances as a source of contrastive demonstrations, 
enabling the $\mathtt{LLM}$ to learn how to resolve conflicts among tools.

%To resolve the conflicts among the tools in $\mathcal{T}_s(r)$, 
%% 
%\method develops a novel tool conflict memory $\mathcal{M}$ 
%% 
%to facilitate comparison between tools 
%% 
%when they disagree on the same reaction.
%% 
%Specifically,
%%
%\method focuses on reactions in the validation set where level-1 tools produce inconsistent predictions.
%%
%For each reaction $r$, the selected tool set $\mathcal{T}_s(r)$ is obtained and partitioned into correctly and incorrectly predicting tools.
%%
%For each correctly predicting tool $t_i^+$, \method constructs a contrastive instance by pairing it with all incorrectly predicting tools $\{t_j^-\}$.
%%
%Each tool is further associated with its pattern in $\mathcal{P}_s(r)$, resulting in a structured instance consisting of the reaction, the correctly predicting tool $t_i^+$, all incorrectly predicting tools $\{t_j^-\}$, and their associated patterns.
%%
%The $\mathtt{LLM}$ is employed to generate a rationale explaining why $t_i^+$ is more suitable for reaction $r$ than $\{t_j^-\}$ based on their associated patterns.
%% 
%The memory $\mathcal{M}$ stores such contrastive reasoning instances,
%%
%which highlight the differences between tools under the same reaction.
%% 
%$\mathcal{M}$ further serves as a source of demonstrations, 
%% 
%enabling the $\mathtt{LLM}$ to learn how to resolve conflicts among tools.

%=========================================================
\paragraph{Memory-augmented Conflict Resolution}
%=========================================================

To address the conflicts among tool predictions during inference time for reaction $r$, 
\method adapts a novel conflict resolution agent via memory-augmented reasoning using \memory. 
Specifically, \method retrieves the top-$K$ most similar reactions to $r$ from \memory based on DRFP representation~\citep{probst2022reaction},
a reaction fingerprint that captures reaction-level structural transformations.
% \xia{Ye, need to explain this and references}.
%
% From the contrastive instances of those similar reactions, 
% \method identifies one pattern for each reaction that is most similar to $\Pbest_t(r)$ ($t\in\SelectedTool(r)$), 
% and thus, one corresponding contrastive instance. %, from their multiple contrastive instances in \memory. 
% %
% Note that such most similar patterns may stem from $\Pbest_{t^+}$ or $\{\Pbest_{t^-}\}$ of the constrastive instances, 
% offering opportunities for \method to consider different tools that have correct or incorrect predictions  
% on similar reactions. 
% %
% The $K$ identified contrastive instances, one for each of the $K$ most similar reactions, are used as few-shot demonstrations, 
% together with $\SelectedTool(r)$ and their associated patterns, to prompt the $\mathtt{LLM}$ to determine the optimal tool 
% for $r$, and thus produce the final prediction. 
% %
% The conflict resolution leverages the reasoning capability of the $\mathtt{LLM}$ 
% by learning from contrastive historical instances whose reactions are similar to $r$ and whose tools exhibit behavioral patterns 
% similar to the tools selected for $r$, 
% %
% including both cases where the tools correctly and incorrectly predicted analogous reactions, 
% thereby providing informative signals for distinguishing reliable tool agreements. 
%
For each of these similar reactions,
\method randomly selects one of its contrastive reasoning instances in \memory,
These instances are used as few-shot demonstrations,
together with the selected tool set $\SelectedTool(r)$ and their associated patterns,
to prompt the $\mathtt{LLM}$ to determine the optimal tool for $r$, and thus produce the final prediction.
The conflict resolution process leverages the reasoning capability of the $\mathtt{LLM}$ by learning from historical contrastive instances of reactions similar to $r$,
thereby providing informative signals for distinguishing reliable tools under conflicts.
\section{Experimental Setting}
\label{sec:setup}
%%%%%%%%%%%%%%%%%%%%%%%%%%%%%%%%%%%

%===================================================
\paragraph{Dataset}
\label{exp:datasets}
%===================================================

\input{tables/data}
We conduct experiments on the \textit{FREA} dataset~\citep{yu2025rxnverif}, 
which is constructed from the U.S. Patent \& Trademark Office (USPTO). 
The dataset contains real-world feasible reactions from USPTO, 
and infeasible reactions generated through 
multiple chemically motivated perturbation strategies validated with expert evaluations,
providing a rigorous testbed for reaction feasibility prediction. 
Specifically, we randomly select 12,000 reactions and split them into a validation set and a test set (Table~\ref{tbl:dataset}).
Since our objective is to leverage multiple tools for reaction feasibility prediction, rather than train individual tools, 
we do not involve model training on this dataset, and thus do not introduce a training split. 
The validation set is used for tool hierarchy construction, tool utility assessment, and tool conflict memory construction,
while the test set is used for evaluating the performance of \method.
% 

%===================================================
\paragraph{Reaction Feasibility Tool Set}
%===================================================
\label{exp:toolset}
The tool set consists of 13 individual tools, including 3 classification-based feaasibility predictors, 2 forward-generative models, and 8 LLM-based feasibility reasoners.
Details on the construction of the tool set are provided in Appendix~\ref{app:tool}.
Importantly,
% draft
we strictly ensure that there is no overlap between the dataset (Table~\ref{tbl:dataset}) and the training data for individual tools (Table~\ref{tbl:app_dataset} in Appendix). 
This setup guarantees that the reported results of our \method framework precisely
reflect the effectiveness of leveraging multiple tools, 
rather than being influenced by data leakage or model memorization.

% \paragraph{Tool Set Construction.}
% \label{subsec:toolset}
% % 
% Given a reaction $r$, 
% %
% \method queries a set of feasibility prediction tools, which can be categorized into three major groups:
% % 
% \begin{itemize}[leftmargin=12pt, itemsep=0pt, topsep=0pt]

%     \item \textit{Classification-based feasibility predictors}: 
%     %
%     These models formulate reaction feasibility as a binary classification task based on reaction representations, 
%     %
%     including DRFP~\citep{probst2022reaction}, Dora\_xgb~\citep{chainani2025dora}, and a BERT-based classifier~\citep{devlin2019bert}.

%     \item \textit{Forward-generation models}: 
%     %
%     These models assess feasibility by predicting plausible products given reactants and checking their consistency with the target products, 
%     %
%     including Molecular Transformer~\citep{schwaller2019molecular} and Chemformer~\citep{irwin2022chemformer}.

%     \item \textit{LLM-based feasibility reasoners}: 
%     %
%     These tools perform feasibility prediction through language-based reasoning, 
%     %
%     including multiple prompting and in-context learning variants built on \texttt{Llama-3.1-8B-Instruct}~\citep{llama3modelcard} and \texttt{T3Q-Qwen-14B}~\citep{t3q_qwen2_5_14b}.

% \end{itemize}
% % 
% Together, these tools form the complete tool set $\mathcal{T^{(\text{all})}} = \{t_i\}$ used in our framework. 
% %
% Detailed descriptions of tool construction and training procedures are provided in Appendix~\ref{app:tool}.

%===================================================
\paragraph{Evaluation Metrics}
%===================================================
%
We evaluate the reaction feasibility prediction performance using Accuracy (ACC), F1 score, and Matthews Correlation Coefficient (MCC)~\citep{chicco2020advantages}.
We report overall accuracy as well as class-wise accuracy for feasible and infeasible reactions.
For F1 score, we report class-wise performance by treating each class (feasible/infeasible) as the positive class in turn.
MCC is further adopted as a balanced metric to provide a comprehensive evaluation.
% under potential class imbalance.

%===================================================
\paragraph{Tool Selection Baselines}
\label{exp:baseline}
%===================================================
%To demonstrate the effectiveness of \method,
%
We compare \method with a diverse set of baselines, grouped into four categories:
\textbf{(1)} \textit{Single-Tool Methods}, which evaluate each individual tool independently;
\textbf{(2)} \textit{Statistical Methods}, which aggregate tool predictions based on predefined statistical rules, including majority voting, weighted voting, random selection, and StaticSel~\citep{britto2014dynamic};
% \ye{StaticSel, where a fixed subset of tools is selected based on validation-set performance, and then combined by majority voting.}
% \xia{is this a name of a well-known method?}
% 
\textbf{(3)} \textit{Dynamic Methods}, which select tools for each instance by estimating tool competence based on its local neighborhoods
% \xia{of what?} 
or input characteristics, including \mbox{DES-KNN} and \mbox{DES-Clustering}~\citep{soares2006using}, \mbox{KNORA-E} and \mbox{KNORA-U}~\citep{ko2008dynamic}, and HarderMoE~\citep{huang2024harder};
and \textbf{(4)} \textit{LLM-based Methods}, which leverage LLMs to assess tool utilities and perform tool selection through semantic reasoning, % rather than explicit statistical estimation, 
including ToolEyes~\citep{ye2025tooleyes} and several closed-source LLMs (e.g., GPT-5.4-mini~\citep{singh2025openai}, DeepSeek-v4-flash~\citep{deepseek2026v4}, and Claude-Sonnet-4.6~\citep{anthropic2026claude46}).
Detailed descriptions of all baselines are provided in Appendix~\ref{app:baseline}.
%
% \hl{We additionally report the theoretical upper bound of the tool set,
% %
% where a prediction is considered correct if any tool produces the correct output.} \xia{no need to 
% discuss here}

%===================================================
\paragraph{Implementation Details}
\label{exp:imp details}
%===================================================

In \method, we adopt \texttt{T3Q-Qwen-14B}~\citep{t3q_qwen2_5_14b} as the backbone LLM,
with default parameter settings and $\mathtt{do\_sample=False}$ to eliminate sampling randomness.
For tool hierarchy construction (Section~\ref{sec:tool_hierarchy}),
we set the proportion $\rho = 25$
and use accuracy on the validation set as the ranking metric.
For pattern extraction (Section~\ref{subsubsec:pattern_extraction}),
we set $M = 100$ and $N \in \{5, 10, 25, 45\}$ to construct diagnostic subsets.
For pattern refinement (Eq.~\ref{eq:p_valid}) and pattern consolidation,
we set $\tau_1 = 1$, $\tau_2 = 1$ and $\tau_3 = 0.5$, respectively.
For pattern-based tool prioritization (Section~\ref{sec:pattern_matching}), 
we set $L = 5$ to control the number of selected tools for prediction.
For memory-augmented conflict resolution (Section~\ref{subsec:conflict}),
we set the number of retrieved similar reactions to $K = 8$.
All experiments are conducted on a Linux server with two Tesla A100 GPUs.

%%%%%%%%%%%%%%%%%%%%%%%%%%%%%%%%%%%
\section{Experimental Results}
%%%%%%%%%%%%%%%%%%%%%%%%%%%%%%%%%%%
\label{sec:Experimental Results}
%====================================================
\subsection{Overall Performance}
%====================================================

\input{tables/main_res}
%\label{tab:main_result}

Table~\ref{tab:main_result} reports the performance of individual tools, 
the theoretical upper bound of leveraging all tools, 
various tool aggregation and tool selection baselines, and our \method framework.
We highlight several key findings from the results as follows:
% \hl{Several notable observations can be drawn:} \xia{Chinglish... }

%-------------------------------------------------------------------------------------------------------------------------
\textbf{\textit{Individual tools exhibit limited performance and substantial variation.}}
%-------------------------------------------------------------------------------------------------------------------------
%
No single tool consistently performs best across all reactions.
For example,
BERT achieves the best overall performance among all individual tools (87.90\% in overall accuracy),
while Chemformer$_{\texttt{Llama}}$ performs best on feasible reactions,
and Chemformer achieves the strongest performance on infeasible reactions,
%
% \hl{Many tools show imbalanced performance across feasible and infeasible reactions.
% %
% For example, some tools achieve strong performance on feasible reactions but perform poorly on infeasible ones (e.g., Chemformer$_{\texttt{Llama}}$), 
% %
% while others exhibit the opposite behavior (e.g., Chemformer, DRFP).} 
% \xia{the point is not to compare between feasible reactions vs infeasible reactions, but to 
% compare across the tools on feasible and infeasible reactiosn, respectly, with the conclusions 
% that they have different performance and no single tool dominates... }
% %
Such diversity leads to a very high theoretical upper bound of best performance of all tools together,
(i.e., the performance obtained when a reaction is considered correctly predicted if any tool predicts it correctly),
reaching 99.90\% overall accuracy,
% (e.g., 99.90\% in overall accuracy),
% \xia{here, explain what you mean by ``theoretical upper bound"}
%
indicating that different tools capture complementary aspects of the task.
This observation validates our motivation for leveraging multiple tools and highlights the significant potential of selecting appropriate tools for each reaction.

%-------------------------------------------------------------------------------------------------------------------------
\textbf{\textit{\method consistently outperforms all baselines with more balanced performance across both feasible and infeasible reactions.}}
%-------------------------------------------------------------------------------------------------------------------------
%
\method achieves the best performance on most metrics, except for accuracy on the infeasible set.
In terms of overall accuracy, it surpasses the strongest individual tool (BERT, 87.90\%) and reaches 91.62\%.
\method also consistently improves over existing tool selection baselines, including
% Compared with the best tool selection baseline, 
HarderMoE (89.50\%), which leverages implicit expert routing to assign more tools to harder cases and fewer to easier ones.
Unlike such implicit routing strategies,
\method explicitly reasons over the utilities of different tools for each reaction
and further resolves conflicts among selected tools,
allowing it to better leverage the complementary strengths of different tools.
In addition, 
\method achieves strong and balanced accuracy performance on both feasible and infeasible reactions,
resulting in superior F1 and MCC scores that reflect a more balanced predictive behavior across different reaction classes.
%
% \hl{{\method} achieves a notable improvement of 2.12\%.} \xia{I would suggest removing this statement - 2.12\% difference is not that huge... } 
% \xia{if you use accuracy as the primary metric, that is fine, but you still need to brielf discuss 
% F1 and MCC here; in later discussions, can focus on accuracy...}
%
% Unlike HarderMoE that relies on implicit optimization for routing decisions,
%
% \method \hl{explicitly assesses tool utilities and selects appropriate tools for each reaction, followed by a conflict resolution agent to resolve potential tool conflicts,} \xia{rephrase, avoid using 
% almost identical language... }
%
% resulting in improved and balanced overall performance, \hl{as well as better interpretability of the decision process.} \xia{is this highlighted in ``Method"? the accuracies cannot demonstrate interpretability... }

%-------------------------------------------------------------------------------------------------------------------------
\textbf{\textit{LLM-based methods achieve competitive performance but often suffer from imbalanced behavior.}}
%-------------------------------------------------------------------------------------------------------------------------
%
Claude-Sonnet-4.6 and ToolEyes rank among the top-performing tool selection methods in terms of overall accuracy.
However,
their performance varies substantially across reaction classes.
% \xia{need to summarize ``competititve performance'' first... }
For example, Claude-Sonnet-4.6 and ToolEyes achieve strong accuracy performance on the infeasible set (96.53\% and 94.77\%, respectively), 
% \xia{up to here, it becomes unclear what these percentages are, accuracy or F-1 or MCC? }
%
but their performance on the feasible set is significantly lower (74.57\% and 87.25\%),
leading to suboptimal overall results.
This further demonstrates the advantage of \method in achieving balanced performance across both feasible and infeasible reactions.
% 
%-------------------------------------------------------------------------------------------------------------------------
\textbf{\textit{Statistical methods fail to outperform the strongest individual tool.}}
%-------------------------------------------------------------------------------------------------------------------------
%
Statistical methods such as majority voting and weighted voting combine predictions using predefined rules without considering input-specific characteristics or tool strengths.
As a result, they are unable to correct systematic errors shared by a majority of tools,
leading to inferior performance.
% 
%-------------------------------------------------------------------------------------------------------------------------
\textbf{\textit{Most dynamic methods show limited improvement over statistical baselines.}}
%-------------------------------------------------------------------------------------------------------------------------
%
KNORA-E achieves performance comparable to majority voting (82.88\%),
suggesting limited benefit from its neighborhood-based selection.
Furthermore, KNORA-U and DES variants underperform compared to majority voting,
indicating that their selection strategies may even introduce suboptimal decisions in this setting.

% \hl{More experiments in Appendix demonstrated that xxx.} \xia{make sure to complete this}
More experiments show that, in {\method}, incorporating more demonstrations improves conflict resolution and leads to more accurate predictions. 
In addition, the tool selection distribution analysis reveals that generally strong \levelOne tools are not frequently selected in conflict resolution,
whereas some specialized \levelTwo tools are chosen substantially more often.
These observations further highlight the importance of tool utility modeling.
More details can be found in Appendix~\ref{app:demonstration} and Appendix~\ref{app:distribution}.

\vspace{-8pt}
%========================================================
\subsection{Ablation Study}
%========================================================
% 
\begin{wrapfigure}{r}{0.4\textwidth}
\vspace{-45pt}
    \centering
    \includegraphics[width=0.68\linewidth]{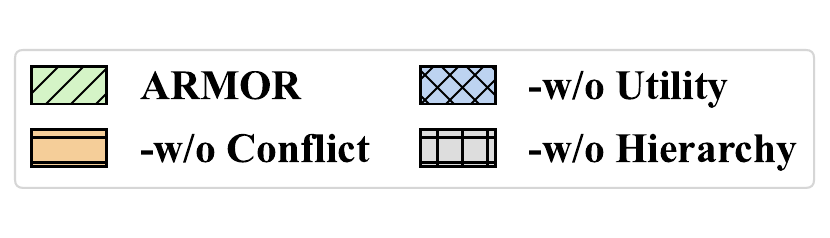}
    \includegraphics[width=0.48\linewidth]{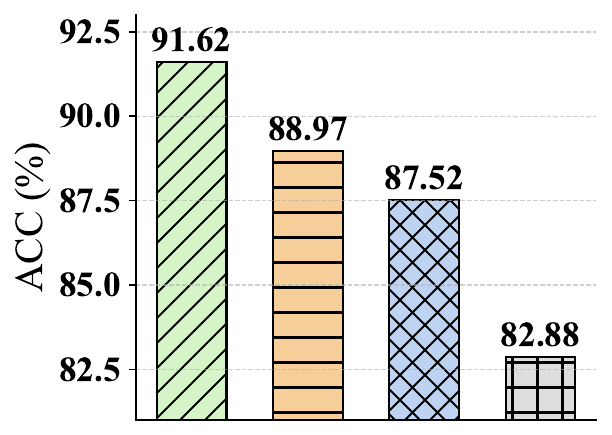}
    \hfill
    \includegraphics[width=0.48\linewidth]{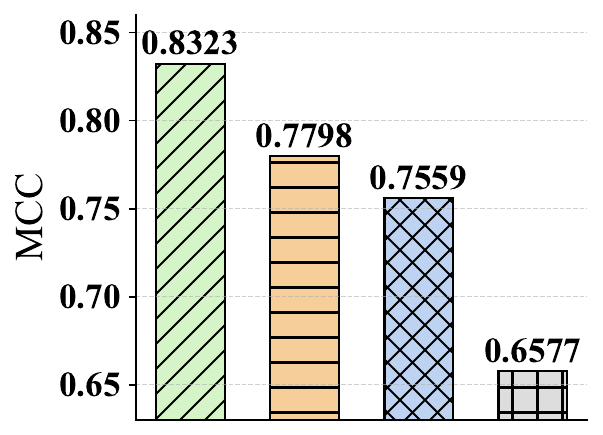}
    % \vspace{-5pt}
    \caption{Ablation study of \method.}
    \vspace{-10pt}
    \label{fig:ablation}
\end{wrapfigure}
In this subsection, we conduct ablation experiments to assess the effectiveness of different components in the \method framework.
We progressively remove the tool conflict resolution module (Section~\ref{subsec:conflict}), utility-aware tool prioritization module (Section~\ref{subsec:cap_tool_sel}) and tool hierarchy construction module (Section~\ref{sec:tool_hierarchy}), 
resulting in three variants:
\textit{-w/o Conflict},
\textit{-w/o Utility},
and \textit{-w/o Hierarchy},
respectively.
% to examine their impact on performance.
% \xia{section references are not correct... }
%
Specifically,
In \textit{-w/o Conflict}, the final prediction degrades to majority voting over the selected tools $\SelectedTool$ in Section~\ref{sec:pattern_matching}.
% \xia{consolidate all these -w/o notations with the above sentence}
%
In \textit{-w/o Utility}, predictions are obtained via majority voting across all tools for reactions that cannot be consistently resolved by \levelOne.
% \hl{level-1 tools} \xia{use notations as in Method},
%
% denoted as \textit{-w/o Utility}.
%
In \textit{-w/o Hierarchy}, the framework is reduced to majority voting across all tools for all reactions.
% Finally, we remove the tool hierarchy, reducing the method to majority voting across all tools for all reactions,
%
% denoted as \textit{-w/o Hierarchy}.
%
% The results are illustrated in Figure~\ref{fig:ablation}.

From the results in Figure~\ref{fig:ablation}, we observe a clear performance degradation as components are progressively removed,
indicating that each component contributes to the overall effectiveness of \method.
In particular, removing tool conflict resolution leads to a noticeable drop,
highlighting the effectiveness of our memory-augmented design in resolving tool conflicts.
Further removing utility-aware tool prioritization causes additional degradation,
demonstrating the necessity of explicitly modeling tool-specific utilities.
The largest performance drop occurs when removing the tool hierarchy,
highlighting its importance in prioritizing top-performing tools for straightforward reactions
and deferring the use of the remaining tools to more challenging cases.

\vspace{-8pt}
%========================================================
\subsection{Performance Gains across Reaction Categories}
%========================================================

% \hl{To better understand where the performance gain comes from, } \xia{this is vague; if with a 
% more intuitive title, can remove this... }
%
In this subsection, we analyze the improvement of \method over the strongest baseline, HarderMoE, across three reaction categories,
including:
(1) reactions where $\levelOne$ produces consistent predictions, denoted as $\levelOne$;
(2) reactions where the selected tools $\SelectedTool$ (Section~\ref{sec:pattern_matching}) produce consistent predictions, denoted as $\SelectedTool$; and 
(3) reactions where $\SelectedTool$ still produce conflicting predictions, denoted as \textit{Conflict}.

% Specifically, all 6,000 testing reactions can fall into 
% % 
% \{\levelOne(consistent), \SelectedTool(consistent), \SelectedTool(conflict)\}.
% % 
% \levelOne(consistent) refers to the reactions where level-1 tools reach consistent predictions.
% % 
% \SelectedTool(consistent) includes the reactions that can be consistently addressed by the selected tools \SelectedTool in Section~\ref{sec:pattern_matching}.
% % 
% \SelectedTool(conflict) represents the reactions there \SelectedTool still produce conflicting predictions.
% %
% \xia{need to simplify these notations, not very intuitive... }

\input{tables/improvement}
%\label{tbl:category_analysis}
%
As shown in Table~\ref{tbl:category_analysis}, 
the improvement is not uniformly distributed across three categories,
but is mainly concentrated on cases where tools produce inconsistent predictions.
Specifically, for reactions where \levelOne reach consistent predictions,
\method achieves only a marginal improvement (+0.26\%),
indicating that 
% \hl{these cases are relatively easy and already well-handled by strong baselines}
these cases can be well handled by strong baselines.
%
% \hl{For reactions that become consistent after tool selection} \xia{unclear what it means... } (\SelectedTool),
For reactions where \SelectedTool produce consistent predictions,
\method achieves a more noticeable improvement (+7.03\%).
For these reactions, \levelOne initially produce inconsistent predictions,
but after pattern-based tool selection, the selected tools (\SelectedTool) reach agreements,
suggesting that \method's utility-aware tool prioritization module effectively identifies appropriate tools for each reaction.
In contrast, for reactions where tools in \SelectedTool still produce conflicting predictions,
which correspond to the most challenging cases,
\method achieves a substantial improvement (+6.89\%).
This demonstrates its effectiveness in resolving persistent tool conflicts under difficult scenarios.
Overall, \method consistently improves over HarderMoE (+2.12\%),
with the gains mainly arising from its ability to identify appropriate tools through tool utility modeling and to resolve remaining conflicts in difficult cases.

%========================================================
\subsection{Case Study}
%========================================================

% \xia{let's discuss this section}
% \begin{figure}[h]
% \centering
% \vspace{-10pt}
%     \begin{subfigure}[t]{0.40\textwidth}
%         \centering
%         \includegraphics[width=\linewidth]{figures/case1.pdf}
%         \caption{Matched tools with consistent predictions.}
%         \label{fig:case1}
%     \end{subfigure}
%     \hspace{0.05\textwidth}
%     \begin{subfigure}[t]{0.44\textwidth}
%         \centering
%         \includegraphics[width=\linewidth]{figures/case2.pdf}
%         \caption{Matched Tools with conflicting predictions.}
%         \label{fig:case2}
%     \end{subfigure}
%     \vspace{-5pt}
%     \caption{Case study of \method framework.}
%     \label{fig:case}
%     \vspace{-10pt}
% \end{figure}

\begin{wrapfigure}{r}{0.43\textwidth}
\vspace{-50pt}
    \centering
    \includegraphics[width=\linewidth]{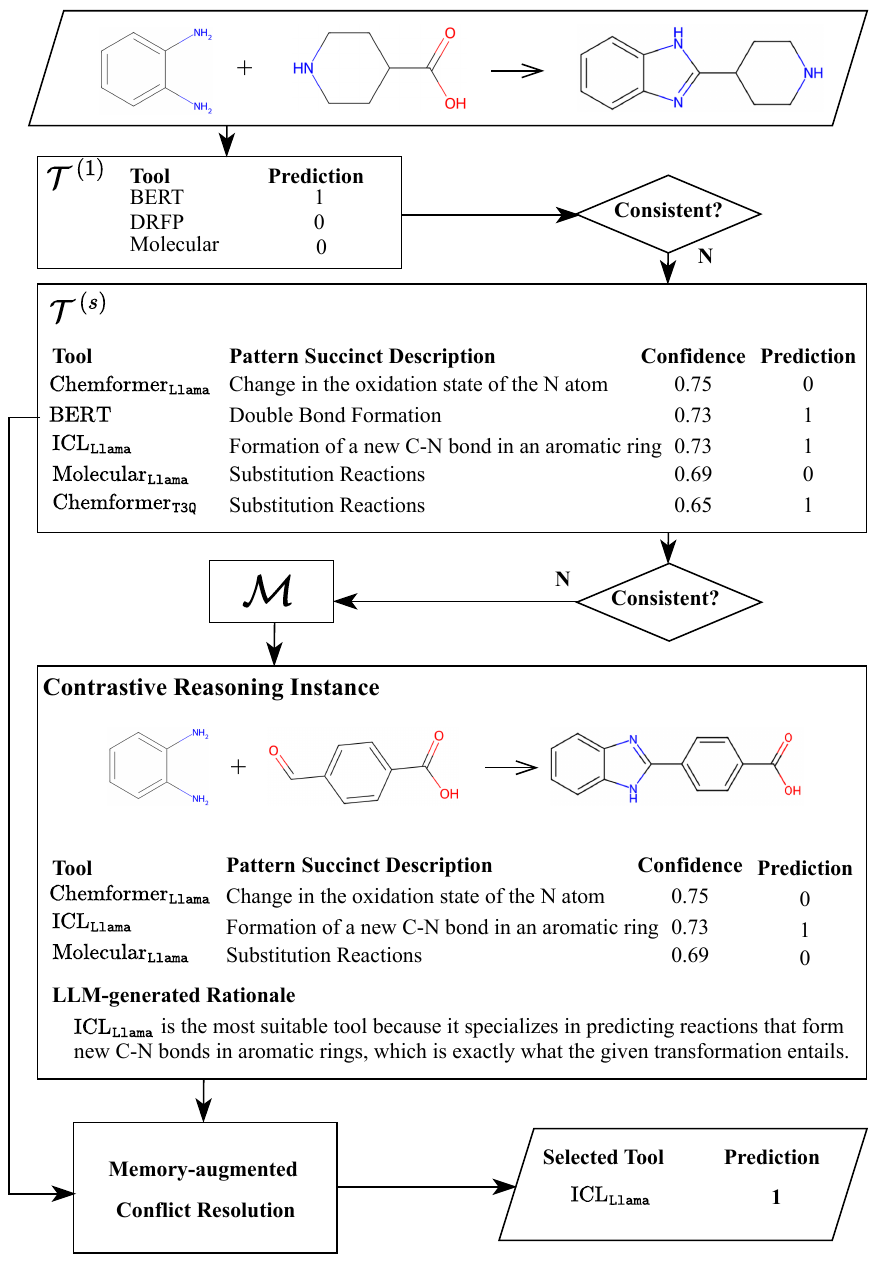}
    \caption{Case study.}
    \vspace{-10pt}
    \label{fig:case}
\end{wrapfigure}
To further illustrate the reasoning process of \method,
we present a representative case study in Figure~\ref{fig:case}.
For the given reaction $r$, \levelOne (BERT, DRFP and Molecular) produce inconsistent predictions,
indicating that generally strong tools alone are insufficient for this reaction.
\method therefore conduct the pattern-based tool selection,
where \SelectedTool are selected together with their associated patterns.
Although these patterns all cover the given reaction,
the selected tools still produce inconsistent predictions.
To resolve the conflict,
\method retrieves the contrastive reasoning instances from the tool conflict memory $\memory$.
One of the retrieved instance is shown in Figure~\ref{fig:case}. It involves a similar reaction to the $r$,
and contains three tools, including one correctly predicting tool (ICL$_{\texttt{Llama}}$) and two incorrectly predicting tools (Chemformer$_{\texttt{Llama}}$, Molecular$_{\texttt{Llama}}$),
with patterns overlapping those of the current reaction $r$.
According to the rationale in this instance,
ICL$_{\texttt{Llama}}$ is preferred because it specializes in reactions involving the formation of new C--N bonds in aromatic rings.
Since the current reaction $r$ exhibits the same transformation pattern,
\method selects ICL$_{\texttt{Llama}}$ and uses its prediction as the final prediction.
This case demonstrates how \method transfers reasoning signals from historical contrastive instances
to identify the most appropriate tool under conflicting predictions, finally deriving the correct prediction.

% To further illustrate the effectiveness of \method, 
% %
% we present two representative case studies in Figure~\ref{fig:case}, 
% %
% demonstrating how \method leverages multiple tools for reaction feasibility prediction 
% % 
% under different conditions.
% % 
% In Figure~\ref{fig:case1}, 
% %
% the matched tools produce consistent predictions. 
% %
% All selected tools correctly predict the reaction as feasible, 
% %
% supported by coherent capability patterns (e.g., \textit{Aromatic ring substitution}). 
% %
% In this case, \method directly accepts the consensus, 
% %
% indicating that pattern-based filtering can effectively identify appropriate tool subsets without additional reasoning.
% % 
% In contrast, Figure~\ref{fig:case2} presents a more challenging scenario, 
% %
% where the matched tools yield conflicting predictions. 
% %
% In this case, 
% %
% \method selects the ICL$_{\texttt{Llama}}$ tool, 
% %
% whose capability pattern (\textit{Formation of a new C--N bond in an aromatic ring}) better matches the reaction. 
% %
% Although the Chemformer$_{\texttt{Llama}}$ tool has the highest confidence, 
% %
% its pattern (\textit{Change in the oxidation state of the N atom}) is less aligned with the reaction, 
% %
% leading to the correct final prediction of \method.
% % 
% More importantly, 
% % 
% through the explicit capability profiling and tool selection process,
% % 
% \method offers a transparent and interpretable decision-making process, 
% % 
% contributing to more accurate and reliable feasibility prediction for each reaction.
% 

%%%%%%%%%%%%%%%%%%%%%%%%%%%%%%%%%%%%%%%%
\section{Conclusions}
%%%%%%%%%%%%%%%%%%%%%%%%%%%%%%%%%%%%%%%%

In this work, we studied how to effectively leverage multiple tools for reaction feasibility prediction, 
where different tools exhibit varying strengths across reactions.
We developed \method, an agent-based framework that integrates tool hierarchy construction, 
utility-aware tool prioritization, and tool conflict resolution 
to guide tool usage and derive the final prediction for each reaction.
By explicitly modeling when different tools succeed, 
\method effectively leverages their strengths, particularly in challenging cases with persistent conflicting predictions.
Extensive experiments demonstrate that \method consistently outperforms strong baselines, highlighting the importance of utility-aware tool prioritization over direct aggregation strategies.
%
% \hl{Beyond performance gains, our framework provides a transparent decision process, offering insights into tool behavior and improving the interpretability.}

\section*{Acknowledgements}
This project was made possible, in part, by support from the National Library of Medicine grant no. R01LM014385 (X.N., D.A.-A.) and the National Science Foundation grant no. 2435819 (X.N., D.A.-A.). Any opinions, findings, and conclusions or recommendations expressed in this manuscript are those of the authors, and do not necessarily reflect the views of the funding agencies.
We thank Reza Averly, Frazier N. Baker, and Benjamin Burns for their help with paper writing. We are grateful for their careful review, constructive feedback, and assistance in improving the clarity and organization of the manuscript.

\bibliography{ref}
\bibliographystyle{ACM-Reference-Format}

%%%%%%%%%%%%%%%%%%%%%%%%%%%%%%%%%%%%%%%%%%%%%%%%%%%%%%%%%%%%

\newpage
\appendix
\renewcommand{\thefigure}{A\arabic{figure}}

\renewcommand{\thetable}{A\arabic{table}}

\setcounter{figure}{0}

\setcounter{table}{0}

\section{Tool Set Construction and Training Details}
\label{app:tool}
\input{tables/tool_training}
%
% In this section,
% %
% we provide additional details on the construction and training of the tools used in \method.
%

\paragraph{Training Data}
The individual tools are trained using the \textit{FREA} dataset~\citep{yu2025rxnverif},
which is derived from the U.S. Patent \& Trademark Office
(USPTO\footnote{\url{https://figshare.com/articles/dataset/Chemical_reactions_from_US_patents_1976-Sep2016_/5104873}}).
The detailed data statistics are summarized in Table~\ref{tbl:app_dataset}.
Following the evaluation protocol in Section~\ref{exp:datasets},
the reactions used for training individual tools are fully disjoint from those used in validation and testing (Table~\ref{tbl:dataset}).
This separation ensures that the performance of \method reflects its ability in leveraging multiple tools,
rather than gains from overlapping training data.

\paragraph{Tool Set}
As introduced in Section~\ref{exp:toolset},
the tool set contains 13 tools, including 3 classification-based feasibility predictors,
2 forward-generative models, and 8 LLM-based feasibility reasoners.
For LLM-based methods, subscripts $_{llama}$ and $_{T3Q}$ indicate the method is based on the \texttt{Llama-3.1-8B-Instruct}~\citep{llama3modelcard} and \texttt{T3Q-Qwen-14B}~\citep{t3q_qwen2_5_14b}, respectively, as the base LLMs. 

% \xia{give an overall summary of this paragraph: we use XXX categories of tools. 
% need to explain subscript (e.g., $_{llama}$ means what?}

\paragraph{(1) Classification-based Feasibility Predictors}
\begin{itemize}[leftmargin=12pt, itemsep=0pt, topsep=0pt]
\item \textbf{BERT}~\citep{devlin2019bert}. 
We fine-tune \texttt{bert-base-uncased} as a sequence classification baseline for reaction feasibility prediction. 
Each reaction is represented as a plain text string by concatenating the reactant and product SMILES with a >> separator (i.e., reactants>>product), 
which is then tokenized using BERT's native WordPiece tokenizer 
and truncated to a maximum length of 128 tokens. 
The model is trained for 50 epochs using AdamW (weight decay = 0.01) with a learning rate of 2e-5, 
a linear warmup over the first 10\% of training steps, and mixed-precision (fp16) training. 
The checkpoint with the highest F1 score on the validation set is selected as the final model.
\item \textbf{DRFP}~\citep{probst2022reaction}. 
Instead of text-based tokenization, we encode each reaction using the Differential Reaction Fingerprint (DRFP), 
which produces a 2048-dimensional binary vector capturing the structural changes between reactants and products. 
The fingerprint is fed into a two-layer MLP (2048$\rightarrow$256$\rightarrow$2) with ReLU activation and dropout (0.1). 
All other training settings follow those of BERT above.
\item \textbf{Dora\_xgb}~\citep{chainani2025dora}. 
We directly deploy the pre-trained DORA-XGB model\footnote{https://github.com/tyo-nu/Dora\_xgb.} without any fine-tuning. 
The model encodes each reaction using ECFP4 fingerprints (2048 bits), 
where molecular fingerprints of individual species are arranged by descending molecular weight to form a fixed-length reaction fingerprint, 
which is then fed into a pre-trained XGBoost classifier to predict feasibility.
\end{itemize}

\paragraph{(2) Forward-Generation Models}
\begin{itemize}[leftmargin=12pt, itemsep=0pt, topsep=0pt]
\item \textbf{Chemformer}~\citep{irwin2022chemformer}. 
We deploy the pre-trained Chemformer forward synthesis model\footnote{https://github.com/MolecularAI/Chemformer.} without fine-tuning. 
Given the reactant SMILES, 
the model performs beam search to generate the top-10 most likely products along with their log-likelihoods. 
A reaction is predicted as feasible if the given product appears among the top-5 candidates 
(after canonicalization and removal of atom mapping).
\item \textbf{Molecular}~\citep{schwaller2019molecular}. 
Following the same inference strategy as Chemformer, 
we deploy the pre-trained Molecular Transformer\footnote{https://github.com/pschwllr/MolecularTransformer.} 
for forward reaction prediction. 
The reactant SMILES is first canonicalized and then tokenized at the atom level before being fed into the model, 
which generates the top-5 predicted products. 
Feasibility is determined by whether the given product is present in the top-5 predictions.
\end{itemize}

\paragraph{(3) LLM-based Feasibility Reasoners}
\begin{itemize}[leftmargin=12pt, itemsep=0pt, topsep=0pt]
\item \textbf{Prompt$_{\texttt{Llama}}$ / Prompt$_{\texttt{T3Q}}$}. 
We directly prompt two instruction-tuned LLMs
--- \texttt{Llama-3.1-8B-Instruct}~\citep{llama3modelcard} and \texttt{T3Q-Qwen-14B}~\citep{t3q_qwen2_5_14b} --- 
for zero-shot feasibility prediction. 
Given only the reaction SMILES (\texttt{reactants>>product}), 
each LLM is asked to evaluate whether the reaction is chemically feasible under plausible laboratory conditions, 
based solely on its internalized chemistry knowledge.
\item \textbf{ICL$_{\texttt{Llama}}$ / ICL$_{\texttt{T3Q}}$}. 
We augment the same two LLMs with retrieved in-context examples. 
For each query reaction, 
we use its DRFP fingerprint~\citep{probst2022reaction} to retrieve the top-3 most similar feasible reactions 
and top-3 most similar infeasible reactions from the training set via FAISS binary index with Hamming distance. 
These six retrieved reactions are provided to the LLM alongside the query, 
and the model is asked to make a feasibility judgment informed by the retrieved evidence.
\item \textbf{Chemformer$_{\texttt{Llama}}$ / Chemformer$_{\texttt{T3Q}}$}. 
We augment the same two LLMs with forward synthesis predictions from Chemformer~\citep{irwin2022chemformer}. 
For each query reaction, 
Chemformer generates the top-10 candidate products via beam search, and the top-5 (after canonicalization and removal of atom mapping) are provided to the LLM as prior evidence. 
The LLM then makes a final feasibility judgment 
by reasoning over whether the given product is consistent with the predicted candidates.
\item \textbf{Molecular$_{\texttt{Llama}}$ / Molecular$_{\texttt{T3Q}}$}. 
Following the same strategy as above, 
we replace Chemformer with the Molecular Transformer~\citep{schwaller2019molecular} 
to generate the top-5 candidate products, 
which are likewise provided to the LLM as contextual evidence for feasibility judgment.
\end{itemize}

\section{Baselines}
\label{app:baseline}

To demonstrate the effectiveness of \method,
we compare it with a range of tool selection baselines,
which can be grouped into four categories:
\begin{itemize}[leftmargin=12pt, itemsep=0pt, topsep=0pt]

    \item \textbf{Single-Tool Methods.} 
    We report the performance of each individual tool in the tool set constructed in Appendix~\ref{app:tool},
    serving as fundamental baselines.

    \item \textbf{Statistical Methods.} 
    StaticSel~\citep{britto2014dynamic} performs majority voting over a fixed subset of tools selected based on development performance.
    In contrast, we also include simple aggregation baselines applied to all tools,
    including majority voting, weighted voting, and random selection.

    \item \textbf{Dynamic Methods.}
    These methods estimate tool competence based on the validation set and select tools dynamically for each instance during testing.
    KNORA-E~\citep{ko2008dynamic} selects tools that correctly classify all samples in the $k$-nearest neighborhood of a test reaction,
    while KNORA-U relaxes this requirement to at least one sample.
    DES~\citep{soares2006using} selects tools by jointly considering tool accuracy and diversity.
    DES-KNN estimates these criteria within the $k$-nearest neighborhood of each test reaction,
    whereas DES-Clustering evaluates them within cluster-defined regions.
    In addition, we include HarderMoE~\citep{huang2024harder}, which performs dynamic expert routing based on input difficulty.

    \item \textbf{LLM-based Methods.}
    ToolEyes~\citep{ye2025tooleyes} employs LLM-based scoring to assess tool utility and select tools for each instance, where the same LLM backbone is adopted as in \method for a fair comparison.
    We also include closed-source LLMs, including GPT-5.4-mini~\citep{singh2025openai}, DeepSeek-v4-flash~\citep{deepseek2026v4} and Claude-Sonnet-4.6~\citep{anthropic2026claude46},
    which are prompted to evaluate tool utilities on the validation set and perform tool selection during testing.

\end{itemize}
% 
% We additionally report the theoretical upper bound of the tool set,
% %
% where a reaction is considered correctly predicted if any tool produces the correct output.

\section{Impact of Demonstrations}
\label{app:demonstration}
\input{tables/demostration}

In this section, we analyze the impact of the number of retrieved demonstrations ($K$) in the memory-augmented conflict resolution part (Section~\ref{subsec:conflict}).
As shown in Table~\ref{tbl:conflict_agent}, 
the overall performance improves as $K$ increases from 0 to 8,
suggesting that incorporating more demonstrations provides more informative evidence for resolving tool conflicts and improving tool selection.
Meanwhile, we also observe a trade-off between feasible and infeasible classes:
smaller $K$ (e.g., $K=2$) achieves the best performance on feasible reactions,
while larger $K$ (e.g., $K=8$) significantly improves the performance on infeasible reactions,
leading to more balanced overall results.
This behavior suggests that when $K$ is small,
the model relies on limited evidence,
which may lead to suboptimal or skewed tool selection decisions.
As $K$ increases, the retrieved instances become more diverse,
providing richer evidence for resolving conflicts among tools, 
thus reducing bias in final predictions.

\section{Tool Selection Distribution}
\label{app:distribution}
\input{tables/distribution}
To better understand the behavior of \method, 
we focus on the 1,103 cases where tools in $\SelectedTool$ produce conflicting predictions.\footnote{Refer to Table~\ref{tbl:category_analysis} for the detailed definition of these 1,103 reactions.}
Table~\ref{tbl:tool_usage} reports the distribution of \method's final selected tools in memory-augmented conflict resolution (Section~\ref{subsec:conflict}).
It further presents the level ($\mathcal{T}^{(1)}$, $\mathcal{T}^{(2)}$) from which each tool comes,
together with its accuracy on the selected conflict cases and the full test set.
%
% The table reports how frequently each tool is selected, together with its performance on the selected cases (Selected ACC) and on the full test set (Overall ACC).
%

$\mathcal{T}^{(1)}$ consists of BERT, Molecular, and DRFP, 
which correspond to the globally strong tools.
However, in the \textit{Conflict} set, only BERT is frequently selected (343 times), while Molecular is rarely selected and DRFP is never selected.
This indicates that tools with strong global performance are not always suitable for resolving these specific cases.
Instead, several tools in $\mathcal{T}^{(2)}$ are selected more frequently.
In particular, Chemformer$_{\texttt{Llama}}$ is selected 490 times, achieving 77.35\% accuracy on the conflict cases where it is selected, compared to 60.55\% on the full test set,
and ICL$_{\texttt{Llama}}$ is selected 162 times, improving accuracy from 55.12\% to 80.25\% on its selected conflict cases.
These substantial improvements demonstrate that \method can effectively identify scenarios where these tools are particularly appropriate,
even though their overall performance is not the strongest.

In addition, some tools exhibit highly specialized behavior.
For example, ICL$_{\texttt{T3Q}}$ is selected only 17 times,
but achieves perfect accuracy (100\%) on these selected cases, compared to 63.38\% on the full test set.
This suggests that certain tools are highly accurate and stable under specific conditions,
highlighting the importance of modeling fine-grained tool utilities.
We also observe that, for some tools, the accuracy on its selected cases is lower than that on the full test set, e.g., BERT, which is the strongest tool in terms of the accuracy on the full test set.
%
% primarily due to BERT, which is the strongest tool in terms of overall accuracy.
%
This indicates that while \method can effectively leverage diverse tools,
there remains room for improvement in better utilizing the globally strongest tool in difficult scenarios.
Finally, a small portion of cases (1.00\%) result in failure,
where no suitable tool can be identified and the prediction degrades to directly asking the LLM.
This also leaves a limited number of instances for further improvement.

%%%%%%%%%%%%%%%%%%%%%%%%%%%%%%%%%%%%%%%%
\section{Limitations \& Discussions}
\label{sec:limitation}
%%%%%%%%%%%%%%%%%%%%%%%%%%%%%%%%%%%%%%%%
%
% While \method demonstrates strong performance in leveraging multiple tools for reaction feasibility prediction, several limitations remain:
% 
In this paper, \method relies on a validation set with ground-truth labels to construct the tool hierarchy and to characterize tool utilities.
While this enables effective modeling of tool strengths,
it assumes access to sufficient labeled data for effective utility characterization,
which may not always be available in practical deployment.
% 
% While this enables effective modeling of tool strengths, \hl{it assumes access to labeled data for utility characterization, which may not always be available in practical deployment.} 
% \xia{I think the point is not ``access to labeled data", but ``access to sufficient labeled data"; let's 
% discuss this}
%
A promising direction for future work is to leverage \method’s own outcomes as weak supervision signals,
enabling continual refinement of tool utilities during inference.
Such an online adaptation mechanism could allow \method to evolve over time and better capture dynamic tool behaviors.

% \xia{is this a limitation?} Although our study focuses on reaction feasibility prediction, the proposed \method framework is not restricted to this specific task.
%
Beyond the current experimental setting,
the proposed \method framework is not restricted to reaction feasibility prediction.
The key idea of explicitly modeling tool utilities and selecting appropriate tools for each input instance can be naturally extended to other domains,
such as AI4Science tasks like molecular retrosynthesis, as well as more general applications such as personalized content generation and recommendation, where different tools exhibit varying performance across inputs.
Future work will explore the generalization of \method to diverse tasks and investigate how to adapt utility modeling to different tool types and data distributions.

\section{Impact Statement}
\label{sec:impact}

This work focuses on improving how multiple tools are leveraged for reaction feasibility prediction, 
which may benefit AI-assisted scientific research and chemical synthesis planning. 
By explicitly modeling tool utilities and providing a transparent decision-making process, 
our framework can enhance the reliability and interpretability of AI systems in scientific workflows, 
potentially facilitating more efficient exploration of chemical reactions and accelerating discovery.

Potential negative impacts are limited but may arise from incorrect predictions. 
Inaccurate feasibility assessments could mislead downstream decision-making in chemical synthesis, 
leading to inefficient use of resources or incorrect experimental directions. 
In addition, over-reliance on automated tool selection systems without expert validation 
may amplify such risks in practical deployment.
To mitigate these risks, our framework emphasizes interpretability and transparency, 
allowing users to better understand and verify tool selection decisions. 
We recommend that such systems be used in a human-in-the-loop manner, 
particularly in high-stakes scientific applications, 
where expert validation remains essential.

\newpage
\input{tables/prompts}

% \section{Technical appendices and supplementary material}
% Technical appendices with additional results, figures, graphs, and proofs may be submitted with the paper submission before the full submission deadline (see above). You can upload a ZIP file for videos or code, but do not upload a separate PDF file for the appendix. There is no page limit for the technical appendices. 

% Note: Think of the appendix as ``optional reading'' for reviewers. The paper must be able to stand alone without the appendix; for example, adding critical experiments that support the main claims to an appendix is inappropriate. 

%%%%%%%%%%%%%%%%%%%%%%%%%%%%%%%%%%%%%%%%%%%%%%%%%%%%%%%%%%%%

% \newpage
% \input{checklist.tex}

\end{document}

%% file: macros.tex
\newcommand{\method}{\mbox{$\mathop{\textsc{ARMOR}}\limits$}\xspace}

\definecolor{darkgreen}{RGB}{0,150,0}

\newcommand{\memory}{\mbox{$\mathop{\mathcal{M}}\limits$}\xspace}
\newcommand{\CTLI}{\mbox{$\mathop{{x}}\limits$}\xspace}

\newcommand{\levelOne}{\mbox{$\mathop{\mathcal{T}^{(1)}}\limits$}\xspace}
\newcommand{\levelTwo}{\mbox{$\mathop{\mathcal{T}^{(2)}}\limits$}\xspace}
\newcommand{\SelectedTool}{\mbox{$\mathop{\mathcal{T}^{(s)}}\limits$}\xspace}

\newcommand{\Rset}{\mbox{$\mathop{\mathcal{R}^{\neg(1)}_v}\limits$}\xspace}
\newcommand{\PSZero}{\mbox{$\mathop{\mathcal{P}_t^{0}}\limits$}\xspace}
\newcommand{\PSOne}{\mbox{$\mathop{\mathcal{P}_t^{\prime}}\limits$}\xspace}
\newcommand{\PSfinal}{\mbox{$\mathop{\mathcal{P}_t}\limits$}\xspace}

%% file: tables/data.tex
\begin{wraptable}{r}{0.4\textwidth}
    \vspace{-45pt}
    \centering
    \caption{Dataset statistics.}
    \label{tbl:dataset}
    \footnotesize
    % \small
    \begin{threeparttable}
    \begin{tabular}{
      @{\hspace{4pt}}l
      @{\hspace{6pt}}r
      @{\hspace{6pt}}r
      @{\hspace{6pt}}r
      @{\hspace{0pt}}
    }
    \toprule
    Split & \#Reactions & \#Feasible & \#Infeasible \\
    \midrule
    Validation  & 6,000 & 3,000 & 3,000 \\
    Test & 6,000 & 3,000 & 3,000 \\
    \bottomrule
    \end{tabular}
    \end{threeparttable}
    \vspace{-10pt}
\end{wraptable}

%% file: tables/main_res.tex
\begin{table*}[t]
    \centering
%    \vspace{-15pt}
    \caption{Overall experimental results. % of our proposed \method method. 
    \underline{Underlined} results indicate the best performance among single-tool methods, 
    % 
%    \textit{italic} values denote the theoretical upper bound, 
    % 
    and \textbf{bold} values highlight the best performance among tool selection methods.
    ``Overall", ``Feasible" and ``Infeasible" represent that the evaluation is on the entire, only feasible, 
    and only infeasible reactions, respectively. 
     $\uparrow$ indicates higher values are better. }
    \vspace{-5pt}
    \label{tab:main_result}
    \begin{threeparttable}
    \renewcommand\arraystretch{0.8}
    {
    {\footnotesize
    \begin{tabular}
    {
      @{\hspace{4pt}}l@{\hspace{4pt}} 
      @{\hspace{4pt}}l@{\hspace{4pt}}
      @{\hspace{5pt}}c@{\hspace{5pt}}
      @{\hspace{5pt}}c@{\hspace{5pt}} 
      @{\hspace{5pt}}c@{\hspace{5pt}}
      @{\hspace{5pt}}c@{\hspace{5pt}}
      @{\hspace{5pt}}c@{\hspace{5pt}}
      @{\hspace{5pt}}c@{\hspace{5pt}}
      @{\hspace{5pt}}c@{\hspace{5pt}}
      @{\hspace{5pt}}c@{\hspace{5pt}}
%      @{\hspace{4pt}}c
    }
    % {c|l|r|r|r|r|r|r }
    \toprule
    \multirow{2}{*}{Category} & \multirow{2}{*}{\centering Methods}&
    \multicolumn{3}{c}{\textbf{ACC} (\%) \ $\uparrow$}  & &
    \multicolumn{2}{l}{\textbf{F-1 Score} (\%) \ $\uparrow$} & & 
    \multirow{2}{*}{\textbf{MCC} $\uparrow$} \\
    % \cmidrule{3-7}
%    \cmidrule(l{2pt}r{25pt}){3-5}
    \cmidrule(r){3-5}
%    \cmidrule(l{2pt}r{25pt}){6-7}
	\cmidrule(r){7-8}
     & & Overall & Feasible & Infeasible & & 
     Feasible & Infeasible & &  \\
    \midrule

    % \multirow{13}{*}{\rotatebox{90}{\centering Single Tool}}&  &  &  &  &  &  &  \\
    \multirow{13}{*}{\makecell[r]{Single-Tool\\Methods }}
    & BERT &  
    \underline{87.90} & 80.30 & 95.50 & & 
    \underline{86.90} & \underline{88.75} & &  
    \underline{0.7669}  \\
    & DRFP & 
    80.62 & 66.13 & 95.10 & & 
    77.33 & 83.07 & & 
    0.6398   \\
    & Dora\_xgb & 
    50.80 & 22.00 & 79.60 & & 
    30.90 & 61.80 & & 
    0.0196  \\
    \cmidrule(r){2-10}
    & Molecular &  
    82.30 & 71.33 & 93.27 & & 
    80.12 & 84.05 & & 
    0.6621 \\
    & Chemformer & 81.88 & 68.23 & \underline{95.53} &  & 79.02 & 84.06 &  & 0.6628   \\
    \cmidrule(r){2-10}
    & Prompting$_{\texttt{T3Q}}$ &  58.07 & 54.30 & 61.83 &  & 56.43 & 59.59 &  & 0.1618  \\
    & Prompting$_{\texttt{Llama}}$ &  54.02 & 23.37 & 84.67 &  & 33.69 & 64.80 &  & 0.1017  \\

    & ICL$_{\texttt{T3Q}}$ &  63.38 & 76.27 & 50.50 &  & 67.56 & 57.97 &  & 0.2770  \\
    & ICL$_{\texttt{Llama}}$ & 55.12 & 89.93 & 20.30 &  & 66.71 & 31.14  & & 0.1426  \\

    & Molecular$_{\texttt{T3Q}}$ &  71.93 & 79.77 & 64.10 &  & 73.97 & 69.55 &  & 0.4442  \\
    & Molecular$_{\texttt{Llama}}$ &  60.93 & 84.97 & 36.90 &  & 68.50 & 48.57 &  & 0.2494  \\
    
    & Chemformer$_{\texttt{T3Q}}$ &  69.98 & 80.33 & 59.63 &  & 72.80 & 66.52 &  & 0.4085  \\
    & Chemformer$_{\texttt{Llama}}$ & 60.55 & \underline{90.70} & 30.40 &  & 69.69 & 43.52 &  & 0.2645  \\
    \midrule
    \multicolumn{2}{c}{\textit{Theoretical Upper Bound}}  & \textit{99.90} & \textit{99.80} & \textit{100.00} &  & \textit{99.90} & \textit{99.90} &  & \textit{0.9980}   \\
    \midrule
    \multirow{4}{*}{\makecell[r]{Statistical\\Methods }}
    & StaticSel & 74.12 & 81.87 & 66.37 &  & 75.98 & 71.94 &  & 0.4882  \\
    & Majority Voting & 82.88 & 82.53 & 83.23 &  & 82.82 & 82.94 &  & 0.6577   \\
    & Weighted Voting &  86.35 & 81.50 & 91.20 &  & 85.65 & 86.98 &  & 0.7304  \\
    & Random Selection &  67.48 & 68.67 & 66.29 &  & 67.86 & 67.09 &  & 0.3497  \\
    % & $\mathcal{T}^{(1)}$- Majority Voting &  86.75 & 75.77 & \textbf{97.73} & 85.12 & 88.06 & 0.7534  \\
    % & $\mathcal{T}^{(1)}$- Weighted Voting &  86.75 & 75.77 & \textbf{97.73} & 85.12 & 88.06 & 0.7534  \\
    % & $\mathcal{T}^{(1)}$- Random Sel &  84.53 & 75.54 & 93.51 & 83.00 & 85.80 & 0.7020  \\
    \midrule
    \multirow{5}{*}{\makecell[r]{Dynamic\\Methods }}
    & DES-KNN & 75.80 & 76.57 & 75.03 &  & 75.98 & 75.61 &  & 0.5161   \\
    & DES-Clustering & 72.20 & 84.90 & 59.50 &  & 75.33 & 68.16 &  & 0.4591   \\
    & KNORA-E & 82.88 & 82.53 & 83.23 &  & 82.82 & 82.94 &  & 0.6577  \\
    & KNORA-U & 81.85 & 80.33 & 83.37 &  & 81.57 & 82.12 &  & 0.6373   \\
    & HarderMoE & 89.50 & 86.03 & 92.97 &  & 89.12 & 89.85 &  & 0.7919 \\
    \midrule
    \multirow{6}{*}{\makecell[r]{LLM-based\\Methods}}
    % & Llama &  &  &  &  &  &  \\
    % & T3Q &  &  &  &  &  &  \\
    % & (\texttt{Llama}) ToolEyes &  &  &  &  &  &  \\
    & ToolEyes & 87.25 & 79.73 & 94.77 &  & 86.21 & 88.14 &  & 0.7536 \\
    & GPT-5.4-mini & 75.98 & 87.93 & 64.03 &  & 78.55 & 72.72 &  & 0.5352  \\
    & DeepSeek-v4-flash & 80.90 & 77.13 & 84.67 &  & 80.15 & 81.59 &  & 0.6198  \\
    & Claude-Sonnet-4.6 & 85.55 & 74.57 & \textbf{96.53} &  & 83.77 & 86.98 &  & 0.7288  \\
    \cmidrule(r){2-10}
    & \method (ours)&  \textbf{91.62} & \textbf{91.57} & 91.67 &  & \textbf{91.61} & \textbf{91.62} &  & \textbf{0.8323}  \\

    % & \textit{\method - plus} &  &  &  &  &  &  \\

    % & \makecell[l]{\textit{\method - ConflictAgent} \\ \textit{- Refinement}} &  &  &  &  &  &  \\
    
    \bottomrule
    \end{tabular}
    }}
    % \begin{tablenotes}
    %     \footnotesize
    %     \item * \textbf{Bold font} represents the optimal result. For baseline methods, we follow their publicly released codes to obtain the results.
    % \end{tablenotes}
    \vspace{-15pt}
    \end{threeparttable}
\end{table*}

%% file: tables/improvement.tex
\begin{wraptable}{r}{0.57\textwidth}
    %\vspace{-25pt}
    \centering
    \caption{Improvement of \method over HarderMoE on the overall accuracy (ACC (\%)).}
    \label{tbl:category_analysis}
    {
    \footnotesize
    \begin{threeparttable}
    \begin{tabular}{
      @{\hspace{4pt}}l
      @{\hspace{6pt}}r
      @{\hspace{6pt}}r
      @{\hspace{8pt}}r
      @{\hspace{4pt}}r
      @{\hspace{6pt}}r
      @{\hspace{4pt}}
    }
    \toprule
    Category & $N$ & Proportion (\%) 
    & \method & HarderMoE & $\Delta$ \\
    \midrule
    \levelOne & 4,328 & 72.13 & 94.41 & 94.15 & +\,0.26 \\
    \SelectedTool & 569 & 9.48 & 93.67 & 86.64 & +\,7.03 \\
    \textit{Conflict} & 1,103 & 18.38 & 79.60 & 72.71 & +\,6.89\\
    \midrule
    Total & 6,000 & 100.00 & 91.62 & 89.50 & +\,2.12 \\
    \bottomrule
    \end{tabular}
    \end{threeparttable}}
    \vspace{-5pt}
\end{wraptable}

%% file: tables/tool_training.tex
\begin{wraptable}{r}{0.4\textwidth}
    \vspace{-5mm}
    \centering
    \caption{Dataset for constructing the tool set.}
    \label{tbl:app_dataset}
    \footnotesize
    % \small
    \begin{threeparttable}
    \begin{tabular}{
      @{\hspace{6pt}}l
      @{\hspace{6pt}}r
      @{\hspace{6pt}}r
      @{\hspace{6pt}}r
      @{\hspace{6pt}}
    }
    \toprule
    Split & \#Reactions & \#Feasible & \#Infeasible \\
    \midrule
    Train  & 200,000 & 40,000 & 160,000 \\
    Validation & 25,000 & 5,000 & 20,000 \\
    \bottomrule
    \end{tabular}
    \end{threeparttable}
    \vspace{-6pt}
\end{wraptable}

%% file: tables/demostration.tex
\begin{wraptable}{r}{0.42\textwidth}
    \vspace{-32pt}
    \centering
    \caption{Impact of the number of demonstrations ($K$) on accuracy (ACC ($\%$)).}
    \label{tbl:conflict_agent}
    \footnotesize
    \begin{threeparttable}
    \begin{tabular}{
      @{\hspace{4pt}}c
      @{\hspace{8pt}}r
      @{\hspace{6pt}}r
      @{\hspace{3pt}}r
      @{\hspace{6pt}}
    }
    \toprule
    % \multirow{2}{*}{\# Demonstration} & \multicolumn{3}{c}{ACC (\%)} \\
    % \cmidrule(lr){2-4}
    \# Demonstration & Overall & Feasible & Infeasible \\
    \midrule
    $K = 0$ & 90.80 & 92.53 & 89.07 \\
    $K = 2$ & 91.12 & \textbf{92.57} & 89.67 \\
    $K = 4$ & 91.30 & 92.50 & 90.10 \\
    $K = 8$ & \textbf{91.62} & 91.57 & \textbf{91.67} \\
    \bottomrule
    \end{tabular}
    \end{threeparttable}
    \vspace{-6pt}
\end{wraptable}

%% file: tables/distribution.tex
\begin{table*}[t]
    \centering
    \begin{threeparttable}
    \caption{Tool selection distribution and performance on \textit{Conflict} set and full test set. 
    Level shows the level ($\mathcal{T}^{(1)}$, $\mathcal{T}^{(2)}$) from which the tool comes.
    \textit{Conflict} set consists of reactions where $\SelectedTool$ produce conflicting predictions.
    % Selected ACC refers to the accuracy on the cases where the tool is selected, while Overall ACC denotes its performance on the full test set. 
    For each tool,
    $\nearrow$ indicates that the accuracy on the selected conflict cases is higher than that on full test set, while $\searrow$ indicates the opposite. “--” denotes that the metric is not applicable.}
    
    \label{tbl:tool_usage}

    \small
    
    \begin{tabular}
    {
      @{\hspace{4pt}}l
      @{\hspace{8pt}}c
      @{\hspace{8pt}}r 
      @{\hspace{12pt}}r 
      @{\hspace{4pt}}r
      @{\hspace{15pt}}r
      @{\hspace{4pt}}r
      @{\hspace{15pt}}c
      @{\hspace{4pt}}
    }
    % {l c c cc cc c}
    \toprule
    \multirow{2}{*}{Tool} 
    & \multirow{2}{*}{Level}
    & \multirow{2}{*}{Proportion (\%)} 
    & \multicolumn{2}{l}{\textit{Conflict} Set} 
    & \multicolumn{2}{l}{Full Test Set} 
    & \multirow{2}{*}{Trend} \\
    
    \cmidrule(l{2pt}r{15pt}){4-5}
    \cmidrule(l{2pt}r{15pt}){6-7}
    
    & & 
    & Count 
    & ACC (\%) 
    & Count 
    & ACC (\%) 
    & \\
    
    \midrule
    Chemformer$_\texttt{Llama}$ & $\mathcal{T}^{(2)}$ & 44.42 & 490 & 77.35 & 6000 & 60.55 & $\nearrow$ \\
    BERT & $\mathcal{T}^{(1)}$ & 31.10 & 343 & 82.51 & 6000 & 87.90 & $\searrow$ \\
    ICL$_\texttt{Llama}$ & $\mathcal{T}^{(2)}$ & 14.69 & 162 & 80.25 & 6000 & 55.12 & $\nearrow$ \\
    Molecular$_\texttt{Llama}$ & $\mathcal{T}^{(2)}$ & 3.26 & 36 & 83.33 & 6000 & 60.93 & $\nearrow$ \\
    Chemformer$_\texttt{T3Q}$ & $\mathcal{T}^{(2)}$ & 2.45 & 27 & 70.37 & 6000 & 69.98 & $\nearrow$ \\
    ICL$_\texttt{T3Q}$ & $\mathcal{T}^{(2)}$ & 1.54 & 17 & 100.00 & 6000 & 63.38 & $\nearrow$ \\
    Molecular$_\texttt{T3Q}$ & $\mathcal{T}^{(2)}$ & 0.73 & 8 & 37.50 & 6000 & 71.93 & $\searrow$ \\
    Chemformer & $\mathcal{T}^{(2)}$ & 0.45 & 5 & 80.00 & 6000 & 81.88 & $\searrow$ \\
    Molecular & $\mathcal{T}^{(1)}$ & 0.36 & 4 & 50.00 & 6000 & 82.30 & $\searrow$ \\
    
    \midrule
    \texttt{Unused tools} & & & & & & & \\
    
    Dora\_xgb & $\mathcal{T}^{(2)}$ & 0.00 & 0 & -- & 6000 & 50.80 & \\
    DRFP & $\mathcal{T}^{(1)}$ & 0.00 & 0 & -- & 6000 & 80.62 & \\
    Prompting$_\texttt{Llama}$ & $\mathcal{T}^{(2)}$ & 0.00 & 0 & -- & 6000 & 54.02 & \\
    Prompting$_\texttt{T3Q}$ & $\mathcal{T}^{(2)}$ & 0.00 & 0 & -- & 6000 & 58.07 & \\
    
    \midrule
    Failure (no tool selected) & / & 1.00 & 11 & -- & -- & -- & \\
    
    \bottomrule
    \end{tabular}
    \end{threeparttable}
    
    \vspace{-6pt}
\end{table*}

%% file: tables/prompts.tex
\section{Prompts}
\label{sec:prompts}
We provide all prompts used by \method here.

\subsection{Pattern Extraction}
\label{sec:prompts_1}
We use the following prompts to extract patterns following Eq.~\ref{eq:extraction}.

\vspace{7pt}
\hrule
\begin{lstlisting}
You are a careful chemistry scientist.
Follow the user's instructions exactly.
You must output ONLY valid JSON text.
Do not use markdown, code blocks, or any text outside the JSON object.

You will be given a single dataset file related to reaction feasibility prediction.

The file is in JSON Lines format, where each line corresponds to one chemical reaction.
Each reaction entry includes:
- idx: an integer identifier for referencing specific reactions;
- reactants: a SMILES string representing the reactant molecules;
- product: a SMILES string representing the reaction product;
- label: a ground-truth binary label indicating whether the reaction is feasible;
- prediction from a single tool, where the prediction may be 0, 1, or NA (missing).

Your task is to analyze predictions from the tool only and provide a concise, pattern-level summary of this tool's behavior.
Focus on recurring trends rather than exhaustive coverage.

You must complete the task in a single response.
Do NOT ask to split the task across messages.
Do NOT request scope confirmation.

Treat NA, None, or missing predictions as WRONG when computing ACC. Please compute exact ACC and extract representative indices programmatically.

Strict rules (must follow):
• Output valid JSON only. Do NOT use code blocks or markdown. Do NOT include any text outside the JSON object.

IMPORTANT PRACTICAL CONSTRAINTS:
• Report reaction patterns the tool is often CORRECT on
• Patterns should capture recurring reaction behaviors and be formulated as decision-relevant categories, such that a given reaction can be judged as either belonging to or not belonging to the pattern based on observable characteristics.

Output the following structure exactly as a single JSON object:

{
  "tool_acc": "xx.xx%",
  "often_correct_on": [
    {
      "name": "Short, human-readable, chemistry-level name.",
      "explanation": "Describe explicit, observable reaction characteristics (e.g., bond changes, functional groups appearing or disappearing, or structural motifs) that enable an LLM to make a YES/NO decision on whether a given reaction belongs to this pattern. Where possible, the explanation should specify both inclusion cues (what must be present) and exclusion cues (what would disqualify a reaction from the pattern).",
      "examples_idx": [0, 0, 0, 0, 0]
    }
  ]
}

Hard constraints:
• Reaction pattern names must be short and human-readable.
• Explanations may include chemical concepts, structural cues, or transformation characteristics, as long as it helps the model make a reliable yes/no decision about whether a given reaction belongs to the pattern. Explanations should prioritize decisiveness and discriminability over abstract or high-level chemical generalization.
• Each examples_idx must contain exactly 5 integers that appear as idx values in the dataset.

Analyze ONLY the dataset provided next. After completing the analysis, internally verify whether the generated reaction patterns and explanations can correctly classify the listed example reactions. If mismatches are found, refine the pattern definitions and explanations to resolve the inconsistencies. Perform this verification and refinement silently, and output ONLY the final, consolidated results that conform to the required output structure. Do NOT output intermediate versions, alternative drafts, or self-correction steps.

Dataset input:
{dataset_text}
\end{lstlisting}
\hrule

\subsection{Pattern Matching}
\label{sec:prompts_2}

The prompt used in pattern refinement, pattern consolidation, and pattern-based tool selection is as follows:

\vspace{9pt}
\hrule
\begin{lstlisting}
You are a careful chemistry reaction-rule evaluator.
Follow the user's instructions exactly.
You must output ONLY valid JSON (no markdown, no extra text).

You will be given ONE reaction rule with name, explanation, and an example.
Your task: for THIS rule ONLY, judge whether THIS example belongs to the rule.

Instructions:
- Judge this example independently.
- Do NOT assume the example is correct.
- Base judgment only on:
  (1) rule name,
  (2) rule explanation,
  (3) example reactants/product SMILES.
- When making the judgment, treat the rule explanation as the primary and authoritative criterion for deciding whether an example belongs to the rule; the rule name serves only as a high-level label.

Output exactly ONE JSON object with format:
{
  "name": <rule name>,
  "idx": <example_idx>,
  "belongs_to_rule": true/false,
  "confidence": "high"|"medium"|"low",
  "reason": "brief explanation (1-2 sentences)"
}

Output requirements:
- Output MUST be valid JSON.
- Output ONE JSON object only.
- Do NOT include any text outside the JSON object.
- Do NOT repeat the input examples verbatim.

Rule input:
{
  "name": {rule_name},
  "explanation": {rule_explanation}
}

Example:
{example_json}
\end{lstlisting}
\hrule

\subsection{Pattern Consolidation}
\label{sec:prompts_3}
The prompt used in pattern consolidation is as follows:
\vspace{7pt}
\hrule
\begin{lstlisting}
You are a careful chemistry scientist.
Follow the user's instructions exactly.
You must output ONLY valid JSON text.
Do not use markdown, code blocks, or any text outside the JSON object.

You are given {n_rules} reaction-pattern rules that all share the same name: "{rule_name}".

These rules describe the same concept but were written separately with slightly different wording.
Your task is to keep exactly ONE best rule and remove all others.

Criteria for the rule to keep:
- Most precise and complete chemistry description
- Clearest language

You MUST remove all rules except the best one. Do NOT keep more than one.

Return ONLY valid JSON (no extra text) in this exact schema:
{
  "keep_index": <integer>,
  "reason": "<1-2 sentences>"
}

Here are the candidates:
{candidates_json}
\end{lstlisting}
\hrule

\subsection{Memory Building}
\label{sec:prompts_4}
We use the following prompt to build the tool conflict memory:
\vspace{7pt}
\hrule
\begin{lstlisting}
You are a careful chemistry scientist.
Follow the user's instructions exactly.
You must output ONLY valid JSON text.
Do not use markdown, code blocks, or any text outside the JSON object.

We are constructing a demonstration for tool selection.

Reaction (SMILES):
- reactants: {reactants}
- product: {product}

{cands_section}

In this demonstration, there is EXACTLY ONE trusted tool:
- trusted tool MUST be: "{gold_tool}"
- the following tools are NOT trusted in this demonstration: {neg_tools_json}

Your job:
Write a concise and stable reasoning trace explaining WHY "{gold_tool}" is chosen over the non-trusted tools.
Base the reasoning primarily on the transformation implied by reactants→product and the rule explanations.
The tool_prediction field can be noisy; do NOT decide purely by majority vote.

Return ONLY valid JSON with this exact schema:
{
  "tool": "{gold_tool}",
  "evidence": ["<2-4 bullet points tied to the transformation and specific explanations>"],
  "elimination": [{"tool": "<non_trusted_tool>", "why_not": "<1 sentence>"} ... up to 3 items],
  "final_reason": "<1-2 sentences summary>"
}

Constraints:
- "tool" must be exactly "{gold_tool}".
- "elimination" tools must be selected from the non-trusted tools provided above and must be among: [{neg_str}].
- Do NOT mention the words: feasible, label, ground truth, training.
\end{lstlisting}
\hrule

\subsection{Tool Selection}
\label{sec:prompts_5}
We use the following tool selection prompt in memory-augmented conflict resolution.
\vspace{7pt}
\hrule
\begin{lstlisting}
You are a careful chemistry scientist.
Follow the user's instructions exactly.
You must output ONLY valid JSON text.
Do not use markdown, code blocks, or any text outside the JSON object.

You are given a chemical reaction:
- reactants (SMILES): {reactants}
- product (SMILES): {product}

{cands_section}

Selection principles:
- Choose the tool whose matched rule best explains the actual chemical transformation.
- Prefer more specific and chemically plausible rule explanations over vague/generic ones.
{conf_hint}{tiebreak}- If multiple candidates match well, consider internal consistency across rules.
- Output "abstain" only if no candidate provides a convincing match.
{demos_section}
Return ONLY valid JSON in this exact schema:
{
  "tool": <one of [{allowed_str}]>,
  "reason": "<1-3 sentences>"
}
\end{lstlisting}
\hrule

%% file: ref.bib
@inproceedings{yang2024subgraph,
  title={Subgraph-based Self-Supervised Learning Framework for Enzymatic Reaction Feasibility Prediction},
  author={Yang, Feng and Liu, Juan and Zhang, Qiang and Yang, Zhihui and Liu, Jianghang and Wu, Guangsheng},
  booktitle={2024 IEEE International Conference on Bioinformatics and Biomedicine (BIBM)},
  pages={779--784},
  year={2024},
  organization={IEEE}
}

@article{llama3modelcard,
  title={Llama 3 Model Card},
  author={AI@Meta},
  year={2024},
  url = {https://github.com/meta-llama/llama3/blob/main/MODEL_CARD.md}
}

@article{chainani2025dora,
  title={DORA-XGB: an improved enzymatic reaction feasibility classifier trained using a novel synthetic data approach},
  author={Chainani, Yash and Ni, Zhuofu and Shebek, Kevin M and Broadbelt, Linda J and Tyo, Keith EJ},
  journal={Molecular Systems Design \& Engineering},
  volume={10},
  number={2},
  pages={129--142},
  year={2025},
  publisher={Royal Society of Chemistry}
}

@article{schwaller2019molecular,
  title={Molecular transformer: a model for uncertainty-calibrated chemical reaction prediction},
  author={Schwaller, Philippe and Laino, Teodoro and Gaudin, Th{\'e}ophile and Bolgar, Peter and Hunter, Christopher A and Bekas, Costas and Lee, Alpha A},
  journal={ACS central science},
  volume={5},
  number={9},
  pages={1572--1583},
  year={2019},
  publisher={ACS Publications}
}

@article{irwin2022chemformer,
  title={Chemformer: a pre-trained transformer for computational chemistry},
  author={Irwin, Ross and Dimitriadis, Spyridon and He, Jiazhen and Bjerrum, Esben Jannik},
  journal={Machine Learning: Science and Technology},
  volume={3},
  number={1},
  pages={015022},
  year={2022},
  publisher={IOP Publishing}
}

@article{zhong2025towards,
  title={Towards global reaction feasibility and robustness prediction with high throughput data and bayesian deep learning},
  author={Zhong, Haowen and Liu, Yilan and Sun, Haibin and Liu, Yuru and Zhang, Rentao and Li, Baochen and Yang, Yi and Huang, Yuqing and Yang, Fei and Mak, Frankie S and others},
  journal={Nature Communications},
  volume={16},
  number={1},
  pages={4522},
  year={2025},
  publisher={Nature Publishing Group UK London}
}

@article{warr2014short,
  title={A short review of chemical reaction database systems, computer-aided synthesis design, reaction prediction and synthetic feasibility},
  author={Warr, Wendy A},
  journal={Molecular informatics},
  volume={33},
  number={6-7},
  pages={469--476},
  year={2014},
  publisher={Wiley Online Library}
}

@article{jorgensen1990cameo,
  title={CAMEO: a program for the logical prediction of the products of organic reactions},
  author={Jorgensen, William L and Laird, Ellen R and Gushurst, Alan J and Fleischer, Jan M and Gothe, Scott A and Helson, Harold E and Paderes, Genevieve D and Sinclair, Shenna},
  journal={Pure and Applied Chemistry},
  volume={62},
  number={10},
  pages={1921--1932},
  year={1990},
  publisher={De Gruyter}
}

@article{aithal2012feasibility,
  title={Feasibility study of the potential use of chemistry based emission predictions for real-time control of modern diesel engines},
  author={Aithal, SM and Upadhyay, D},
  journal={Applied Energy},
  volume={91},
  number={1},
  pages={475--482},
  year={2012},
  publisher={Elsevier}
}

@article{fooshee2018deep,
  title={Deep learning for chemical reaction prediction},
  author={Fooshee, David and Mood, Aaron and Gutman, Eugene and Tavakoli, Mohammadamin and Urban, Gregor and Liu, Frances and Huynh, Nancy and Van Vranken, David and Baldi, Pierre},
  journal={Molecular Systems Design \& Engineering},
  volume={3},
  number={3},
  pages={442--452},
  year={2018},
  publisher={Royal Society of Chemistry}
}

@article{park2022machine,
  title={Machine learning applications for chemical reactions},
  author={Park, Sanggil and Han, Herim and Kim, Hyungjun and Choi, Sunghwan},
  journal={Chemistry--An Asian Journal},
  volume={17},
  number={14},
  pages={e202200203},
  year={2022},
  publisher={Wiley Online Library}
}

@inproceedings{murakumo2023llm,
  title={LLM drug discovery challenge: A contest as a feasibility study on the utilization of large language models in medicinal chemistry},
  author={Murakumo, Kusuri and Yoshikawa, Naruki and Rikimaru, Kentaro and Nakamura, Shogo and Furui, Kairi and Suzuki, Takamasa and Yamasaki, Hiroyuki and Nishigaya, Yuki and Takagi, Yuzo and Ohue, Masahito},
  booktitle={AI for Accelerated Materials Design-NeurIPS 2023 Workshop},
  year={2023}
}

@article{qin2024tool,
  title={Tool learning with foundation models},
  author={Qin, Yujia and Hu, Shengding and Lin, Yankai and Chen, Weize and Ding, Ning and Cui, Ganqu and Zeng, Zheni and Zhou, Xuanhe and Huang, Yufei and Xiao, Chaojun and others},
  journal={ACM Computing Surveys},
  volume={57},
  number={4},
  pages={1--40},
  year={2024},
  publisher={ACM New York, NY}
}

@article{qu2025tool,
  title={Tool learning with large language models: A survey},
  author={Qu, Changle and Dai, Sunhao and Wei, Xiaochi and Cai, Hengyi and Wang, Shuaiqiang and Yin, Dawei and Xu, Jun and Wen, Ji-Rong},
  journal={Frontiers of Computer Science},
  volume={19},
  number={8},
  pages={198343},
  year={2025},
  publisher={Springer}
}

@inproceedings{qintoolllm,
  title={ToolLLM: Facilitating Large Language Models to Master 16000+ Real-world APIs},
  author={Qin, Yujia and Liang, Shihao and Ye, Yining and Zhu, Kunlun and Yan, Lan and Lu, Yaxi and Lin, Yankai and Cong, Xin and Tang, Xiangru and Qian, Bill and others},
  booktitle={The Twelfth International Conference on Learning Representations},
  year={2024}
}

@article{probst2022reaction,
  title={Reaction classification and yield prediction using the differential reaction fingerprint DRFP},
  author={Probst, Daniel and Schwaller, Philippe and Reymond, Jean-Louis},
  journal={Digital discovery},
  volume={1},
  number={2},
  pages={91--97},
  year={2022},
  publisher={Royal Society of Chemistry}
}

@inproceedings{devlin2019bert,
  title={Bert: Pre-training of deep bidirectional transformers for language understanding},
  author={Devlin, Jacob and Chang, Ming-Wei and Lee, Kenton and Toutanova, Kristina},
  booktitle={Proceedings of the 2019 conference of the North American chapter of the association for computational linguistics: human language technologies, volume 1 (long and short papers)},
  pages={4171--4186},
  year={2019}
}

@misc{t3q_qwen2_5_14b,
  title = {T3Q-Qwen2.5-14B-v1.0-e3},
  author = {JungZoona},
  year = {2025},
  howpublished = {\url{https://huggingface.co/JungZoona/T3Q-qwen2.5-14b-v1.0-e3}},
  note = {Accessed: 2026-04-15}
}

@article{kojima2022large,
  title={Large language models are zero-shot reasoners},
  author={Kojima, Takeshi and Gu, Shixiang Shane and Reid, Machel and Matsuo, Yutaka and Iwasawa, Yusuke},
  journal={Advances in neural information processing systems},
  volume={35},
  pages={22199--22213},
  year={2022}
}

@article{brown2020language,
  title={Language models are few-shot learners},
  author={Brown, Tom and Mann, Benjamin and Ryder, Nick and Subbiah, Melanie and Kaplan, Jared D and Dhariwal, Prafulla and Neelakantan, Arvind and Shyam, Pranav and Sastry, Girish and Askell, Amanda and others},
  journal={Advances in neural information processing systems},
  volume={33},
  pages={1877--1901},
  year={2020}
}

@article{krishnan2026biomodelsrag,
  title={BioModelsRAG: A Biological Modeling Assistant Using RAG (Retrieval Augmented Generation)},
  author={Krishnan, Bhavyahshree Navaneetha and Heydarabadipour, Adel and Sauro, Herbert},
  journal={arXiv preprint arXiv:2601.22684},
  year={2026}
}

@inproceedings{rubin2022learning,
  title={Learning to retrieve prompts for in-context learning},
  author={Rubin, Ohad and Herzig, Jonathan and Berant, Jonathan},
  booktitle={Proceedings of the 2022 conference of the North American chapter of the association for computational linguistics: human language technologies},
  pages={2655--2671},
  year={2022}
}

@article{cruz2020deslib,
  title={DESlib: A Dynamic ensemble selection library in Python},
  author={Cruz, Rafael MO and Hafemann, Luiz G and Sabourin, Robert and Cavalcanti, George DC},
  journal={Journal of Machine Learning Research},
  volume={21},
  number={8},
  pages={1--5},
  year={2020}
}

@article{rokach2010ensemble,
  title={Ensemble-based classifiers},
  author={Rokach, Lior},
  journal={Artificial intelligence review},
  volume={33},
  number={1},
  pages={1--39},
  year={2010},
  publisher={Springer}
}

@inproceedings{dietterich2000ensemble,
  title={Ensemble methods in machine learning},
  author={Dietterich, Thomas G},
  booktitle={International workshop on multiple classifier systems},
  pages={1--15},
  year={2000},
  organization={Springer}
}

@article{britto2014dynamic,
  title={Dynamic selection of classifiers—a comprehensive review},
  author={Britto Jr, Alceu S and Sabourin, Robert and Oliveira, Luiz ES},
  journal={Pattern recognition},
  volume={47},
  number={11},
  pages={3665--3680},
  year={2014},
  publisher={Elsevier}
}

@article{ko2008dynamic,
  title={From dynamic classifier selection to dynamic ensemble selection},
  author={Ko, Albert HR and Sabourin, Robert and Britto Jr, Alceu Souza},
  journal={Pattern recognition},
  volume={41},
  number={5},
  pages={1718--1731},
  year={2008},
  publisher={Elsevier}
}

@article{woloszynski2012measure,
  title={A measure of competence based on random classification for dynamic ensemble selection},
  author={Woloszynski, Tomasz and Kurzynski, Marek and Podsiadlo, Pawel and Stachowiak, Gwidon W},
  journal={Information Fusion},
  volume={13},
  number={3},
  pages={207--213},
  year={2012},
  publisher={Elsevier}
}

@inproceedings{huang2024harder,
  title={Harder task needs more experts: Dynamic routing in MoE models},
  author={Huang, Quzhe and An, Zhenwei and Zhuang, Nan and Tao, Mingxu and Zhang, Chen and Jin, Yang and Xu, Kun and Chen, Liwei and Huang, Songfang and Feng, Yansong},
  booktitle={Proceedings of the 62nd Annual Meeting of the Association for Computational Linguistics (Volume 1: Long Papers)},
  pages={12883--12895},
  year={2024}
}

@inproceedings{shazeer2017outrageously,
  title={Outrageously Large Neural Networks: The Sparsely-Gated Mixture-of-Experts Layer},
  author={Shazeer, Noam and Mirhoseini, Azalia and Maziarz, Krzysztof and Davis, Andy and Le, Quoc and Hinton, Geoffrey and Dean, Jeff},
  booktitle={International Conference on Learning Representations},
  year={2017}
}

@article{chicco2020advantages,
  title={The advantages of the Matthews correlation coefficient (MCC) over F1 score and accuracy in binary classification evaluation},
  author={Chicco, Davide and Jurman, Giuseppe},
  journal={BMC genomics},
  volume={21},
  number={1},
  pages={6},
  year={2020},
  publisher={Springer}
}

@inproceedings{soares2006using,
  title={Using accuracy and diversity to select classifiers to build ensembles},
  author={Soares, Rodrigo GF and Santana, Alixandre and Canuto, Anne MP and de Souto, Marc{\'\i}lio Carlos Pereira},
  booktitle={The 2006 IEEE international joint conference on neural network proceedings},
  pages={1310--1316},
  year={2006},
  organization={IEEE}
}

@inproceedings{ye2025tooleyes,
  title={Tooleyes: Fine-grained evaluation for tool learning capabilities of large language models in real-world scenarios},
  author={Ye, Junjie and Li, Guanyu and Gao, Songyang and Huang, Caishuang and Wu, Yilong and Li, Sixian and Fan, Xiaoran and Dou, Shihan and Ji, Tao and Zhang, Qi and others},
  booktitle={Proceedings of the 31st international conference on computational linguistics},
  pages={156--187},
  year={2025}
}

@misc{yu2025rxnverif,
  author       = {Yu, Botao and Adu-Ampratwum, Daniel and Baker, Frazier N. and Zhou, Bo and Chen, Ziru and Averly, Reza and Liu, Ye and Gao, Wenhao and Ning, Xia and Sun, Huan},
  title        = {{FREA}: Benchmarking Chemical Reaction Feasibility with Systematic Negatives},
  year         = {2026},
  publisher    = {GitHub},
  howpublished = {\url{https://github.com/OSU-NLP-Group/FREA}},
}

@article{singh2025openai,
  title={Openai gpt-5 system card},
  author={Singh, Aaditya and Fry, Adam and Perelman, Adam and Tart, Adam and Ganesh, Adi and El-Kishky, Ahmed and McLaughlin, Aidan and Low, Aiden and Ostrow, AJ and Ananthram, Akhila and others},
  journal={arXiv preprint arXiv:2601.03267},
  year={2025}
}

@misc{anthropic2026claude46,
  title={Claude Sonnet 4.6 System Card},
  author={Anthropic},
  year={2026},
  howpublished={https://www.anthropic.com},
  note={\url{https://www-cdn.anthropic.com/78073f739564e986ff3e28522761a7a0b4484f84.pdf}}
}

@misc{deepseek2026v4,
  title        = {DeepSeek-V4: Towards Highly Efficient Million-Token Context Intelligence},
  author       = {DeepSeek-AI},
  year         = {2026},
  howpublished = {\url{https://huggingface.co/deepseek-ai/DeepSeek-V4-Pro/blob/main/DeepSeek_V4.pdf}},
  note         = {Accessed: 2026-04-29}
}
